\documentclass[conference]{IEEEtran}
 \IEEEoverridecommandlockouts
\usepackage{adjustbox}
\usepackage{subcaption}
\usepackage{amsmath}
\usepackage{float}
\begin{document}




\title{Training Data Set Assessment for Decision-Making in a Multiagent Landmine Detection Platform\thanks{ This research received funds from De Montfort University,  Colciencias [grant 647, 2014]  and Pontificia Universidad Javeriana [grant VRI-05,2017].}}

\author{
\IEEEauthorblockN{Johana Florez-Lozano}
\IEEEauthorblockA{
\textit{Pontificia Universidad Javeriana}\\
Bogota, Colombia \\
johana.florez@javeriana.edu.co}
\and
\IEEEauthorblockN{Fabio Caraffini}
\IEEEauthorblockA{
\textit{De Montfort University}\\
Leicester, UK \\
fabio.caraffini@dmu.ac.uk}
\and
\IEEEauthorblockN{Carlos Parra}
\IEEEauthorblockA{
\textit{Pontificia Universidad Javeriana}\\
Bogota, Colombia \\
carlos.parra@javeriana.edu.co}
\and
\IEEEauthorblockN{Mario Gongora}
 \IEEEauthorblockA{
\textit{De Montfort University}\\
Leicester, UK\\
mgongora@dmu.ac.uk}
}

\maketitle

\begin{abstract}

Real-world problems such as landmine detection require multiple sources of information to reduce the uncertainty of decision-making. A novel approach to solve these problems includes distributed systems, as presented in this work based on hardware and software multi-agent systems. To achieve a high rate of landmine detection, we evaluate the performance of a trained system over the distribution of samples between training and validation sets. Additionally, a general explanation of the data set is provided, presenting the samples gathered by a cooperative multi-agent system developed for detecting improvised explosive devices. The results show that input samples affect the performance of the output decisions, and a decision-making system can be less sensitive to sensor noise with intelligent systems obtained from a diverse and suitably organised training set.

\end{abstract}


\section{Introduction}

Land mines are considered the most problematic kind of unexploited ordnance (UXO) since they can remain active and dangerous for many years after being concealed. Moreover, they affect the development of communities limiting the use of land where they are concealed and in general are in detriment to health. Generally, there are two types of mines relating to their construction. Military land mines are designed under a set of specifications to target humans or vehicles as tanks, as they pass over or near them \cite{Habib2007}. Handmade devices called improvised explosive devices (IEDs) with no recorded construction specifications, very cheap, easy to make, to deploy and more difficult to be detected with the techniques currently available in the literature (an exhaustive review is available in \cite{Bhope2015}). This means that a general scheme does not exits, and in turn, that a wide range of features must be taken into consideration to design a general detection method. 

Sensing land characteristic and detecting buried objects is very complex and no one sensor or method can achieve this optimally due to several issues, such as the presence of obstacles, the variation in temperature and humidity at different times, the type of ground, among others; this limits the desired (high) detection rate \cite{Takahashi2011}. Hence, the idea of investigating Artificial Intelligence (AI) based data fusion methods arose, with the aim of aggregating heterogeneous pieces of information onto a robust distributed Decision-Making (DM) system implemented on a robotic platform. 

The work proposed in this article is part of the evaluation of the performance of a cooperative and distributed decision-making system; whereby the distribution of tests samples among two sets, training and validation data is done such that it maximised the limited data available. The details and results of the tests performed are presented as follows:

\begin{itemize}
    \item section \ref{sec:previiusResearch} provides a brief background and context to this research by discussing limitations of the current state of the art, revising literature in land-mine detection; 
    \item section \ref{sec:MAPS} introduces and describes the hardware and software system which acquired the samples evaluated;
    \item section \ref{sec:expsetup} gives details on the experimental phase and presents some results;
    \item section \ref{sec:conclusions} draws the conclusions of this work and presents plans.
\end{itemize}

\section{Current Research} \label{sec:previiusResearch}

Recent research in land mine detection aims at increasing accuracy in detecting mines, e.g., a machine-learning sensor fusions method is proposed in \cite{Prado2017a} to lower the false-positive rate, and moves towards the automation of risky manual work done by operators, who still may have to use dog units \cite{Prada2015} or terrain-scanning devices \cite{besaw2015deep}. In addition, there is much research around robotic platforms for searching the land for safer mine detection, more recently aiming to take the acquisition away from the hazardous terrain, e.g., the systems in \cite{Colorado2017} and \cite{Dogaru2019} use a drone to scan a test terrain with a Ground Penetrating Radar (GPR).

However, there is a necessity to reach a higher degree of integration between multiple hardware and software developments, i.e., robotic platforms, sensors, decision-making algorithms and AI algorithms to produce intelligent behaviour and enhance performance. Little has been done to include multiple sensors and search explicitly for improvised explosive devices.

As part of the solution, this piece of research presents the assessment of the training and validation data distribution according to the performance of a cooperative and distributed decision-making system (CoD2M). This decision-making system executes in a decentralised and self-organising manner, called here Multi-Agent Perception System (MAPS), inspired by successful studies as those in \cite{Mathews2013,Ye2016,bib:Zuo18,bib:lian2019} to detect buried objects in the ground. The complete data set that each agent used in this study can be found in the repository \cite{florez2019}.

\section{Multi-Agent System} \label{sec:MAPS}

Our data set is generated by a multi-agent system described by two parts: hardware and software.

\begin{figure}[ht!]
\centering
	\includegraphics[width=0.95\columnwidth]{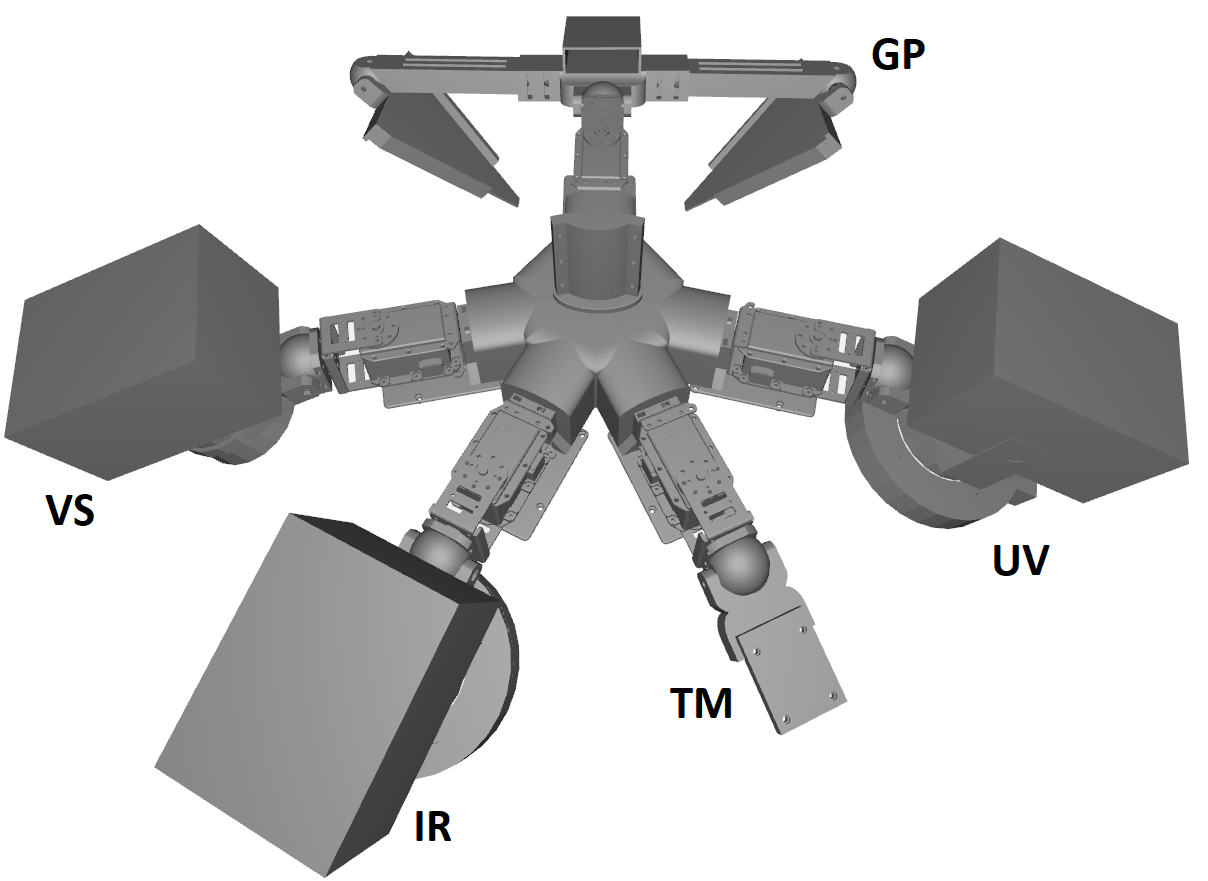}
	\caption{\label{fig:MAPSs}The proposed Multi-Agent System (MAPS) and its sensors. More images are available in \cite{florez2019}}.
\end{figure}

The hardware part consists of a multi-agent perception system (MAPS) with five physical agents (see Figure \ref{fig:MAPSs}). Each agent is equipped with a specific sensor, i.e, a visual spectrum camera (VS), a near-infrared camera (IR), a near-ultraviolet camera (UV), a beam thermal sensor (TM), or a light GPR (GP); each with a Single Board Computer (SBC), and a servomotor as an actuator. The sensors in the MAPS were selected from the performance and advantages of a set of commonly used sensors in past studies and other plausible ones as listed in the review articles \cite{bib:robledo2009,Florez2016}.

The software is divided among the agents of the system where every agent has a cooperative and distributed decision-making system (CoD2M), to make them capable of performing decisions individually. The DM processes which take place are 1) the next acquisition position for the sensor ($\alpha$); 2) a decision whether or not a detected buried object is an IED by using the local information from the integrated sensor ($\beta$); and collaboratively to decide if 3) a decision whether or not a detected buried object is an IED with the support received from the other agents ($\Omega$).

It is important to highlight that the collaborative decision ($\Omega$) can be considered as the definitive output of an agent as it plays the most important role in terms of IED detection. This is because once the five $\Omega$ values are available (i.e. one per agent), they are evaluated and the best one (e.g. the greatest) is the one marking the difference between an IED/non-IED classification (based on $\Omega$ being $<0.5$ or $\geq0.5$ respectively).

In this study, several AI and machine-learning algorithms were chosen from the literature for implementing the three DM processes. These algorithms are divided into three sets, intelligent, statistic aggregation and voting mode. Also, the tests made have all possible configurations among the selected decision-making methods, on each agent. This allowed making a comparative analysis in which a high number of DM systems were tested to recommend the most suitable distribution of training and validation samples to achieve the most robust decision-making system in terms of environmental conditions. 

For the intelligent methods, we used an evolutionary methodology with two test Evolutionary Algorithms (EA). This was due to the possibility to encode individuals as neurons, activation functions, weighted connections, or even fuzzy rules \cite{Baron2001,Stanley2019}, to then manipulate them and generate new neural or fuzzy structures.

The first intelligent decision-making method (IDMM) applied for the $\alpha$, $\beta$ and $\Omega$ decisions is a feed-forward artificial neural networks (ffANN) constructed using a neuroevolution technique. In this work, to train and implement the networks we used the NeuroEvolution of Augmenting Topologies (NEAT) algorithm originally designed in \cite{Stanley2004}. This method is currently used for evolving problem-specific neural structures that would be difficult to devise otherwise \cite{SOLTOGGIO201848}. It contains a GA customised to efficiently optimise parameters like the number of neurons, layers, activation function, connection weights, etc., of an artificial neural network according to the strategy explained in \cite{Stanley2004,Stanley2019}).

A second IDMM is a EA variant designed and used in this study to evolve a fuzzy decision support system (FDSS).  We use a variant of the discrete GA in \cite{Baron2001}, whose individuals (i.e. candidate solutions) encode triangular membership functions. The proposed variant is almost identical to the one in \cite{Baron2001} and employs the same two mutation operators as well as the same three basic crossover strategies, but differs from the original implementation as an additional fourth two-points crossover is used to add more diversity in the generated offspring solutions. This simple modification was tested empirically and adopted as it resulted in better performance. This second method is used for the $\alpha$ and $\Omega$ decisions.

For the case of the aggregation methods, it was only used to reach the $\Omega$ decision. There are three mathematical aggregations (AG) operators maximum (max), average (avg) and median (mdn). They are applied into a set B which contains the five $\beta$ values from each agent, to return:
\begin{equation}
    \lambda_{AG_i} =\left\{\begin{array}{lc}
        \text{max}\left(\text{B}\right) & \text{if\quad}i=1\\
        \text{avg}\left(\text{B}\right) & \text{if\quad}i=2\\
        \text{mdn}\left(\text{B}\right) & \text{if\quad}i=3\\
    \end{array}\right.\quad i=1,2,3.
\end{equation}

And finally, for the last method, we tested the voting mode $\lambda_{\text{VP}}$. This is an approach to aggregate all five $\beta$ values. An agent has a positive vote (i.e. has detected an IED) when its local decision value $\beta\geq 0.5$ and a quorum ( $\lambda_{\text{VP}} = 1$) is reached with at least three votes (since there are five agents in total), otherwise, $\lambda_{\text{VP}} = 0$. Similar to the previous case, this method is only used to reach the outcome given by $\Omega$ decision.

Table \ref{tab:abbreviations} has a summary of the abbreviations of the decision-making methods evaluated and two methods which give the angle value for the next acquisition position ($\alpha$) without an algorithmic process, a fixed point (FP) and a random value (RND). These abbreviations are useful to show the best results listed in figures \ref{fig:ROC_T} and \ref{fig:ROC_V}.

\begin{table}[H]
    \begin{adjustbox}{width=\columnwidth,center=\columnwidth}
     \renewcommand{\arraystretch}{1.5}
    \begin{tabular}{|c||c|c|c|c|c|c|c|c|c|c|}
    \hline\hline
         Method & ffANN & FDSS & FP & RND & $\lambda_{VP}$ & $\lambda_{AG_1}$ & $\lambda_{AG_2}$ & $\lambda_{AG_3}$ \\
         \hline
         Abbreviation & N & F & P & R & V & M & $\overline{\text{B}}$ & $\dot{\text{B}}$ \\
         \hline\hline
    \end{tabular}
    \end{adjustbox}
    \caption{Abbreviations list for the decision-making methods.}
    \label{tab:abbreviations}
\end{table}

To use, test or benchmark against this system, all source code is available online in \cite{codefigshare} as well as in the project's collection \cite{florez2019} along with the data sets.

\section{Experiments and Benchmarking}\label{sec:expsetup}

We selected the required metrics to evaluate the performance of CoD2M-MAPS and its detection accuracy as explained below. It can be noted that almost in the totality of the cases $\Omega$ and $\beta$ are random values distributed normally in the $[0,1]$ range. However, these results can be divided into two sets, to binarize the results, employing a threshold value. Since, there are only two classes, IED and no-IED. 

The most usual metrics for binary classifiers are derived from the confusion matrix \cite{Fawcett2006}, which can be used to annotate the occurrences of a True Positive (TP), a True Negative (TN), a False Positive (FP) and a False Negative (FN) outcome as shown in table \ref{tab:Confusion-Matrix.-TP:}. This leads to the calculation of \\1) the True Positive Rate (TPR):
\begin{equation}\label{eq:TPR}
\text{TPR}=\frac{\text{TP}}{\text{TP}+\text{FN}}=\frac{\text{TP}}{\text{P}};
\end{equation}
2) the False Positive Rate (FPR):
\begin{equation}
\text{FPR}=\frac{\text{FP}}{\text{FP}+\text{TN}}=\frac{\text{FP}}{\text{N}},\label{eq:FPR}
\end{equation}
3) and, the accuracy (ACC):
\begin{equation}
\text{ACC}=\frac{\text{TP}+\text{TN}}{\text{P}+\text{N}}.\label{eq:ACC}
\end{equation}

\begin{table}[H]
\renewcommand{\arraystretch}{1.5}
\begin{centering}
\begin{tabular}{cc|c|c|}
 & \multicolumn{1}{c}{} & \multicolumn{2}{c}{True }\tabularnewline
 & \multicolumn{1}{c}{} & \multicolumn{2}{c}{condition}\tabularnewline
\cline{3-4} 
 &  & Negative &  Positive\tabularnewline
\cline{2-4} 
\multicolumn{1}{c|}{Decision} & Negative & TN & FN\tabularnewline
\cline{2-4} 
\multicolumn{1}{c|}{selected} &  Positive & FP & TP\tabularnewline
\cline{2-4} 
\end{tabular}
\par\end{centering}
\caption{\label{tab:Confusion-Matrix.-TP:} Confusion matrix. In this specific study, the term ``positive'' refers to the detection of an IED, while ``negative'' of a non-IED.}
\end{table}

In addition, a Receiver Operating Characteristic (ROC) curve \cite{Fawcett2006} can display FPR (usually reported on x-axis) and TPR (usually reported on the y-axis) per threshold value. A point in the ROC space is interpreted as a better classification than another one if it is closer to $(0,1)$, as it means that there are only TP and no FP occurrences. Any classifier with a ROC curve or values below the line $x=y$ displays worse performances than a decision made with a random guess. For this reason, the area under the ROC curve (AUC) evaluation metric is commonly used with ROC to compare between classifiers, as its value for a random guess is known and equal to $0.5$. Thus, a classifier with AUC$\leq 0.5$ should never be taken under consideration, while AUC values greater than $0.5$ indicate suitable classifiers. In this light, the best $\beta$, and the best and worst $\Omega$ outcome decisions obtained were plotted in the ROC space in figures \ref{fig:ROC_T} and \ref{fig:ROC_V}, per each distribution sample case. 

As a complement of these binary classification metrics is the Root Mean Square Error (RMSE), defined as
\begin{equation}
\text{RMSE}=\sqrt{\frac{1}{S}\sum_{i=1}^{S}\left(x_{i}-\widehat{x}_{i}\right)^{2}},\label{eq:RMSE}
\end{equation}
\noindent with $x_{i}$ being the expected outcome, $\widehat{x_{i}}$ the predicted output and S the training data-set size. This metric is selected since it is normalised by the sample size S thus allowing for a fair comparison between different DM methods. 

\subsection{Experimental set-up}

The Cartesian robot visible in \cite{florez2019} was built to move CoD2M-MAPS across a test terrain of size is $670\times1100$ mm with a fixed scanning step of $50$ mm. To simulate a real scenario, three mock improvised explosive devices (IEDs) were buried at locations $(550,250)$, $(350,600)$ and $(500,850)$. This test terrain was kept as intact a possible to avoid undesired debris, e.g. city garbage, to contaminate the tests. As in real improvised land mines, plastic bottles and PVC pipes were used for the body of the mock mine while syringes and wooden clothespins were used for simulating the detonation system. The land was left undisturbed time enough for vegetation to grow back and look even again.

\begin{figure}[ht!]
\centering
	\includegraphics[width=1\columnwidth]{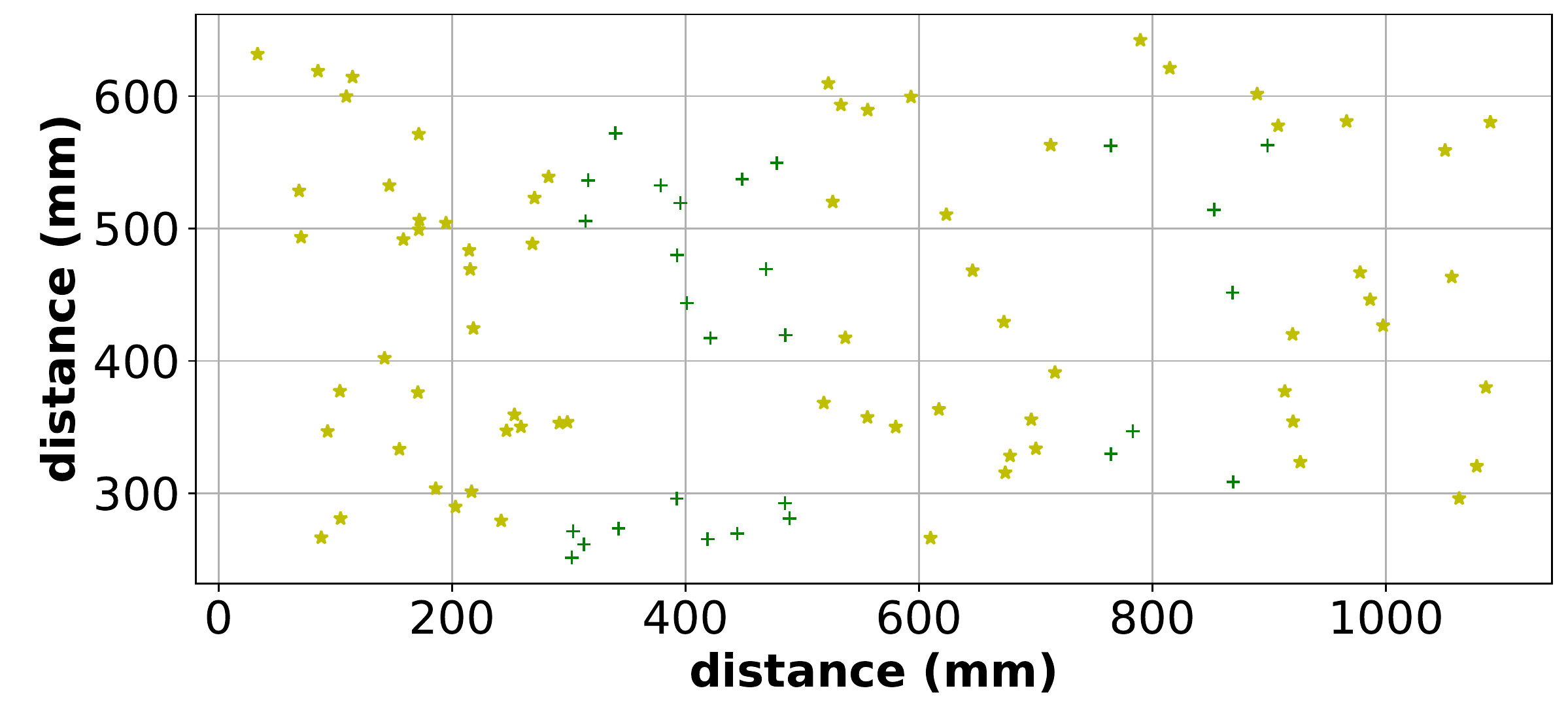}
	\caption{\label{fig:distri}Training (green crosses) and validation (yellow crosses) samples distribution.}
\end{figure}

We collected $100$ samples at fixed positions for two consecutive days for both training and validation (positions are presented in figure \ref{fig:distri}). As expected in real life outdoor scenarios, each day was under different conditions of illumination and relative humidity. For day one (30) sensor acquisitions were made during the morning with a dry day. Samples from day two (31) were during the afternoon and were under drizzle conditions.

Moreover, an assessment of the decision-making system performance according to the distribution of the samples was made for three cases. The first case ($C_1$) uses the samples from day one as the training data set and the models obtained are validated with the samples from day two. For the second case ($C_2$), the samples from day two are used for the training set and the samples from day one for the validation stage. And finally, the third case ($C_3$) divides the samples of both days into two regions indicated in figure \ref{fig:distri} by the colour of the crosses. These regions guaranteed that both sets included data with IED and non-IED samples.

\subsection{Performance evaluation}\label{SCM}

To execute an evaluation of the decision-making systems obtained, the results of the training and validation stages are compared per each case. 

Tables \ref{tab:TR_LOC} and \ref{tab:TR_COOP} belong to the results of the training stage, and tables \ref{tab:VA_LOC} and \ref{tab:VA_COOP} to the validation stage, local and cooperative models respectively. Each table presents the best accuracy model per agent and per case. Also, these tables show the results of each model according to three metrics: the accuracy (ACC -- see equation \ref{eq:ACC}), the root mean square error  (RMSE -- see equation \ref{eq:RMSE}), and the AUC. Besides, the best result per column (per metric) are shown in black bold text. 

\begin{table}
\centering
\begin{subtable}{1\columnwidth}
\begin{adjustbox}{width=\columnwidth,center=\columnwidth}
\begin{centering}
\renewcommand{\arraystretch}{1.5}
\begin{tabular}{|c|c|c|c|c|c|c|c|c|c|}
\cline{2-10} 
 \multicolumn{1}{c|}{} & \multicolumn{3}{c|}{$C_1$} & \multicolumn{3}{c|}{$C_2$} & \multicolumn{3}{c|}{$C_3$}\tabularnewline
\hline 
SEN & ACC & RMSE & AUC & ACC & RMSE & AUC & ACC & RMSE & AUC\tabularnewline
\hline 
\hline 
VS  & 0,690 & \textbf{0,465} & 0,775 & 0,640 & 0,488 & 0,826 & 0,678 & 0,544 & \textbf{0,906}\tabularnewline
\hline 
IR & \textbf{0,730} & 0,498 & \textbf{0,846} & 0,600 & 0,572 & 0,771 & \textbf{0,785} & \textbf{0,411} & 0,818\tabularnewline
\hline 
UV & 0,660 & 0,480 & 0,743 & \textbf{0,820} & \textbf{0,373} & \textbf{0,891} & 0,535 & 0,549 & 0,853\tabularnewline
\hline 
TM & 0,630 & 0,527 & 0,527 & 0,490 & 0,533 & 0,535 & 0,464 & 0,524 & 0,742\tabularnewline
\hline 
GP & 0,340 & 0,504 & 0,500 & 0,590 & 0,453 & 0,676 & 0,785 & 0,418 & 0,830\tabularnewline
\hline 
\end{tabular}
\par\end{centering}
\end{adjustbox}
\caption{\label{tab:TR_LOC}Local decision-making ($\beta$)}
\vspace{0.2cm}
\end{subtable}

\begin{subtable}{1\columnwidth}
\begin{adjustbox}{width=\columnwidth,center=\columnwidth}
\begin{centering}
\renewcommand{\arraystretch}{1.5}
\begin{tabular}{|c|c|c|c|c|c|c|c|c|c|}
\cline{2-10} 
\multicolumn{1}{c|}{} & \multicolumn{3}{c|}{$C_1$} & \multicolumn{3}{c|}{$C_2$} & \multicolumn{3}{c|}{$C_3$}\tabularnewline
\hline 
SEN & ACC & RMSE & AUC & ACC & RMSE & AUC & ACC & RMSE & AUC\tabularnewline
\hline 
\hline 
VS & 0,780 & 0,394 & 0,812 & 0,690 & 0,416 & 0,855 & 0,678 & 0,477 & 0,912\tabularnewline
\hline 
IR & 0,740 & \textbf{0,383} & 0,868 & 0,690 & 0,419 & 0,849 & 0,678 & 0,494 & 0,760\tabularnewline
\hline 
UV & 0,760 & 0,428 & 0,803 & 0,770 & 0,447 & 0,685 & 0,571 & 0,539 & 0,947\tabularnewline
\hline 
TM & \textbf{0,810} & 0,416 & 0,765 & \textbf{0,850} & \textbf{0,331} & \textbf{0,901} & \textbf{0,785} & 0,454 & 0,631\tabularnewline
\hline 
GP & 0,740 & 0,412 & \textbf{0,876} & 0,680 & 0,442 & 0,862 & 0,678 & \textbf{0,398} & \textbf{1,000}\tabularnewline
\hline 
\end{tabular}\end{centering}
\end{adjustbox}
\caption{\label{tab:TR_COOP}Cooperative decision-making ($\Omega$)}
\end{subtable}
\caption{Metrics of the best accuracy result per sensor and per case for the training data set} \label{tab:Train}
\end{table}
\begin{table}
\centering
\begin{subtable}{1\columnwidth}
\begin{adjustbox}{width=\columnwidth,center=\columnwidth}
\begin{centering}
\renewcommand{\arraystretch}{1.5}
\begin{tabular}{|c|c|c|c|c|c|c|c|c|c|}
\cline{2-10} 
\multicolumn{1}{c|}{} & \multicolumn{3}{c|}{$C_1$} & \multicolumn{3}{c|}{$C_2$} & \multicolumn{3}{c|}{$C_3$}\tabularnewline
\hline 
SEN & ACC & RMSE & AUC & ACC & RMSE & AUC & ACC & RMSE & AUC\tabularnewline
\hline 
\hline 
VS & 0,660 & 0,583 & 0,220 & 0,670 & 0,572 & 0,528 & \textbf{0,763} & 0,477 & 0,777\tabularnewline
\hline 
IR & \textbf{0,660} & 0,582 & 0,783 & 0,340 & 0,808 & \textbf{0,842} & 0,722 & \textbf{0,435} & 0,809\tabularnewline
\hline 
UV & 0,660 & 0,583 & \textbf{0,825} & 0,380 & 0,704 & 0,777 & 0,708 & 0,448 & \textbf{0,866}\tabularnewline
\hline 
TM & 0,340 & 0,537 & 0,510 & \textbf{0,670} & \textbf{0,469} & 0,512 & 0,375 & 0,560 & 0,499\tabularnewline
\hline 
GP & 0,420 & \textbf{0,494} & 0,566 & 0,390 & 0,512 & 0,672 & 0,458 & 0,550 & 0,552\tabularnewline
\hline 
\end{tabular}
\par\end{centering}
\end{adjustbox}
\caption{\label{tab:VA_LOC}Local decision-making ($\beta$)}
\vspace{0.2cm}
\end{subtable}

\begin{subtable}{1\columnwidth}
\begin{adjustbox}{width=\columnwidth,center=\columnwidth}
\begin{centering}
\renewcommand{\arraystretch}{1.5}
\begin{tabular}{|c|c|c|c|c|c|c|c|c|c|}
\cline{2-10} 
\multicolumn{1}{c|}{} & \multicolumn{3}{c|}{$C_1$} & \multicolumn{3}{c|}{$C_2$} & \multicolumn{3}{c|}{$C_3$}\tabularnewline
\hline
SEN & ACC & RMSE & AUC & ACC & RMSE & AUC & ACC & RMSE & AUC\tabularnewline
\hline 
\hline 
VS & 0,660 & 0,499 & 0,522 & 0,670 & 0,491 & 0,500 & 0,639 & 0,463 & 0,757\tabularnewline
\hline 
IR & 0,660 & 0,529 & 0,524 & 0,470 & 0,485 & \textbf{0,716} & \textbf{0,777} & \textbf{0,427} & 0,809\tabularnewline
\hline 
UV & \textbf{0,690} & \textbf{0,488} & 0,759 & 0,670 & 0,490 & 0,500 & 0,625 & 0,437 & 0,815\tabularnewline
\hline 
TM & 0,660 & 0,490 & \textbf{0,804} & 0,670 & 0,477 & 0,500 & 0,736 & 0,513 & 0,678\tabularnewline
\hline 
GP & 0,660 & 0,547 & 0,780 & \textbf{0,670} & \textbf{0,463} & 0,610 & 0,638 & 0,422 & \textbf{0,824}\tabularnewline
\hline 
\end{tabular}\end{centering}
\end{adjustbox}
\caption{\label{tab:VA_COOP}Cooperative decision-making ($\Omega$)}
\end{subtable}
\caption{Metrics of the best accuracy result per sensor and per case for the validation data set} \label{tab:Validate}
\end{table}

\begin{figure*}
\centering
	\begin{subfigure}[b]{0.32\textwidth}
	\centering
		\includegraphics[width=1\textwidth,height=0.7\textwidth]{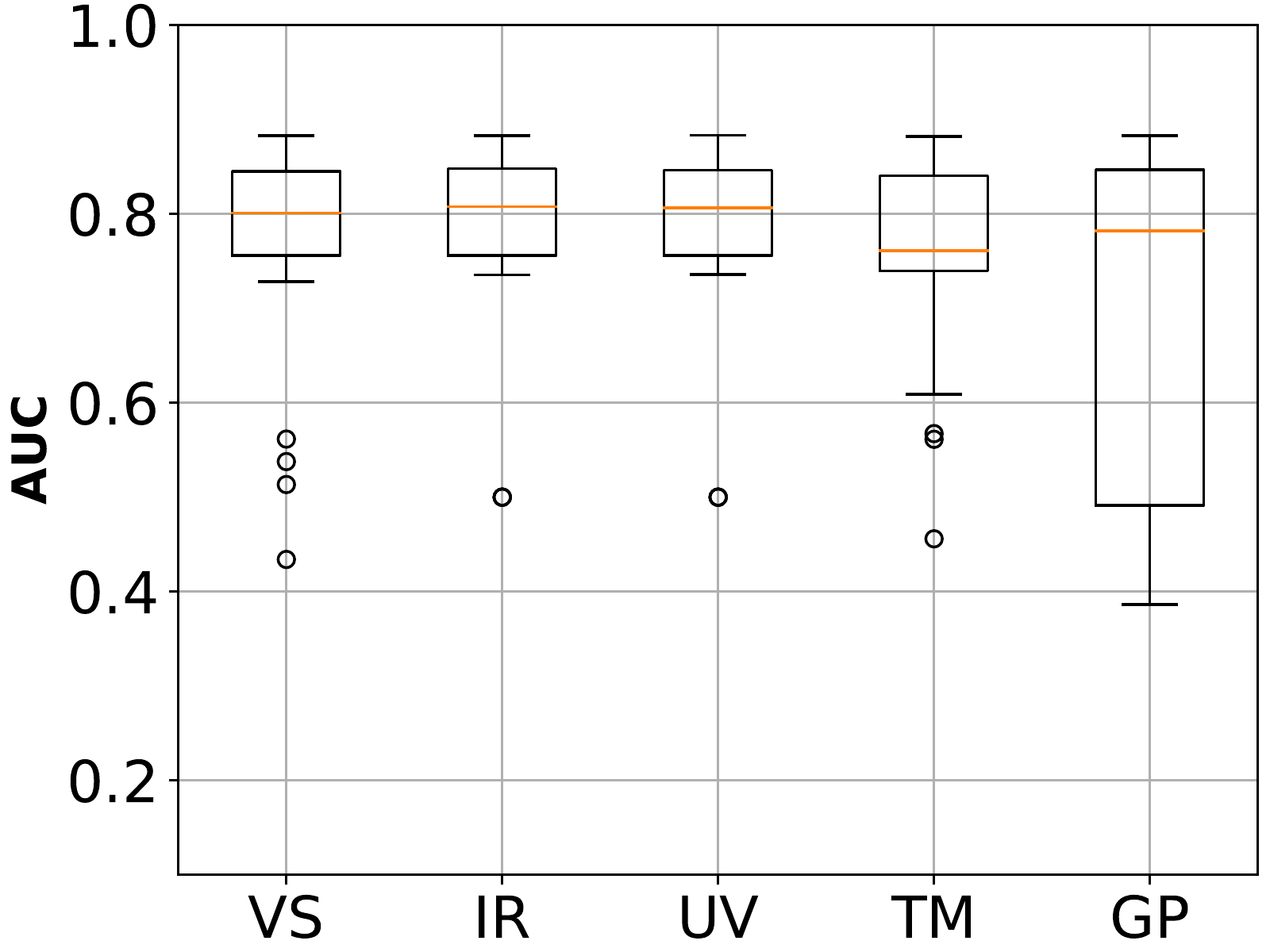}
		\caption{Case 1. Training results}\label{fig:ATCC1}
			\vspace{0.1cm}
	\end{subfigure}
    \hfill
    \hfill\begin{subfigure}[b]{0.32\textwidth}
	\centering
		\includegraphics[width=1\textwidth,height=0.7\textwidth]{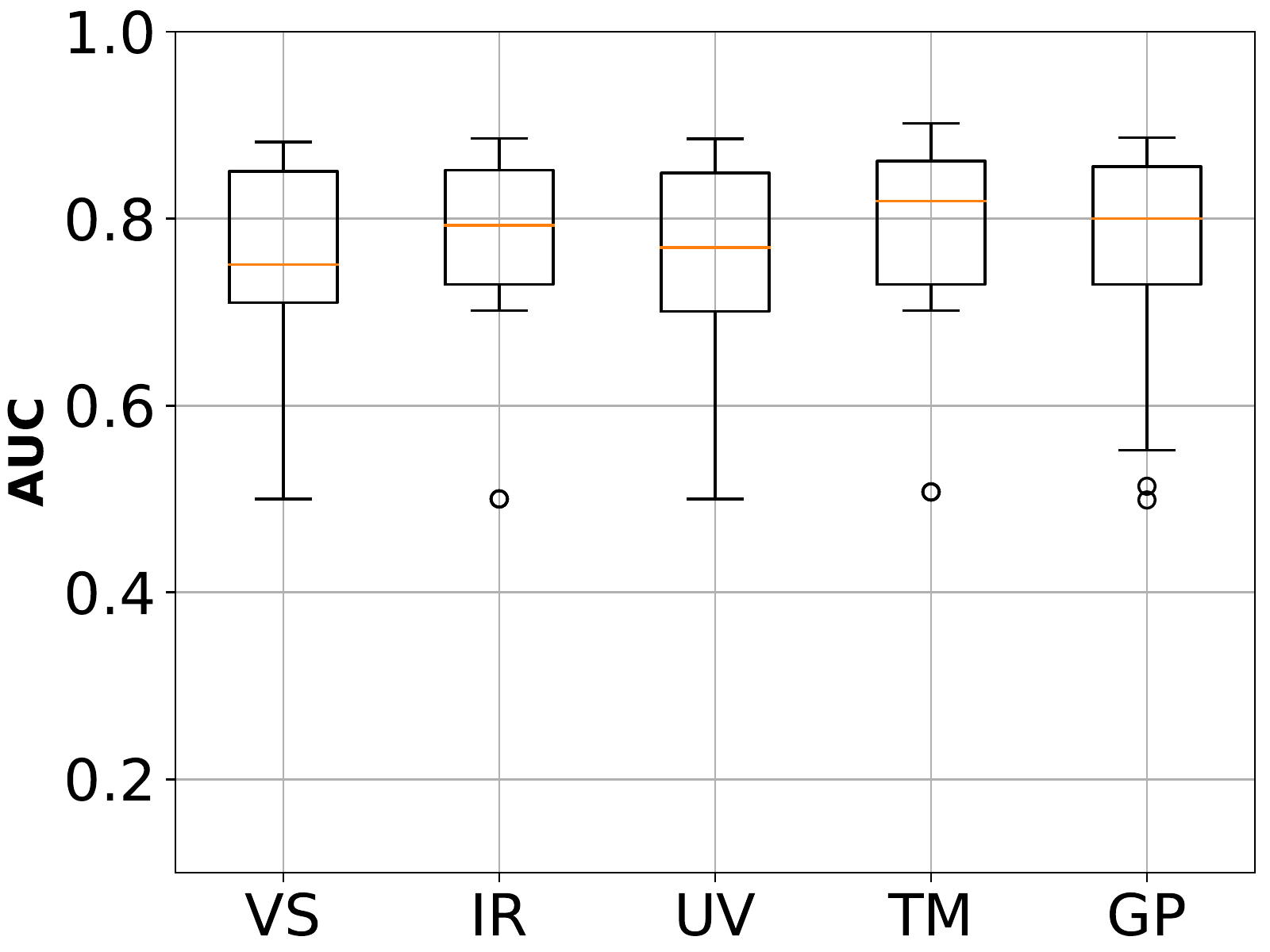}
		\caption{Case 2. Training results}\label{fig:ATCC2}
			\vspace{0.1cm}
	\end{subfigure}
    \hfill
    \begin{subfigure}[b]{0.32\textwidth}
	\centering
		\includegraphics[width=1\textwidth,height=0.7\textwidth]{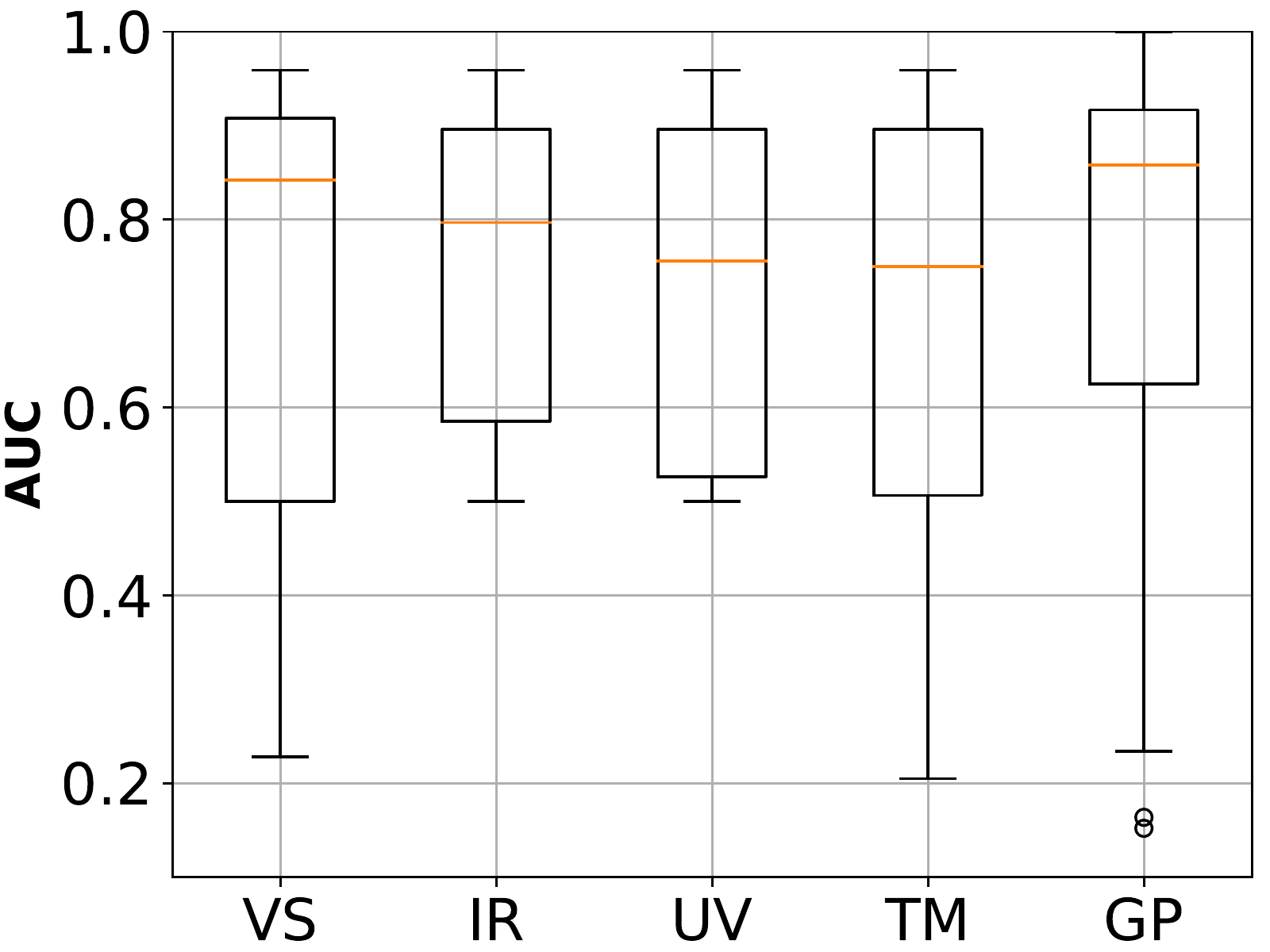}
		\caption{Case 3. Training results}\label{fig:ATCC3}
			\vspace{0.1cm}
	\end{subfigure}
    \hfill
    \begin{subfigure}[b]{0.32\textwidth}
	\centering
		\includegraphics[width=1\textwidth,height=0.7\textwidth]{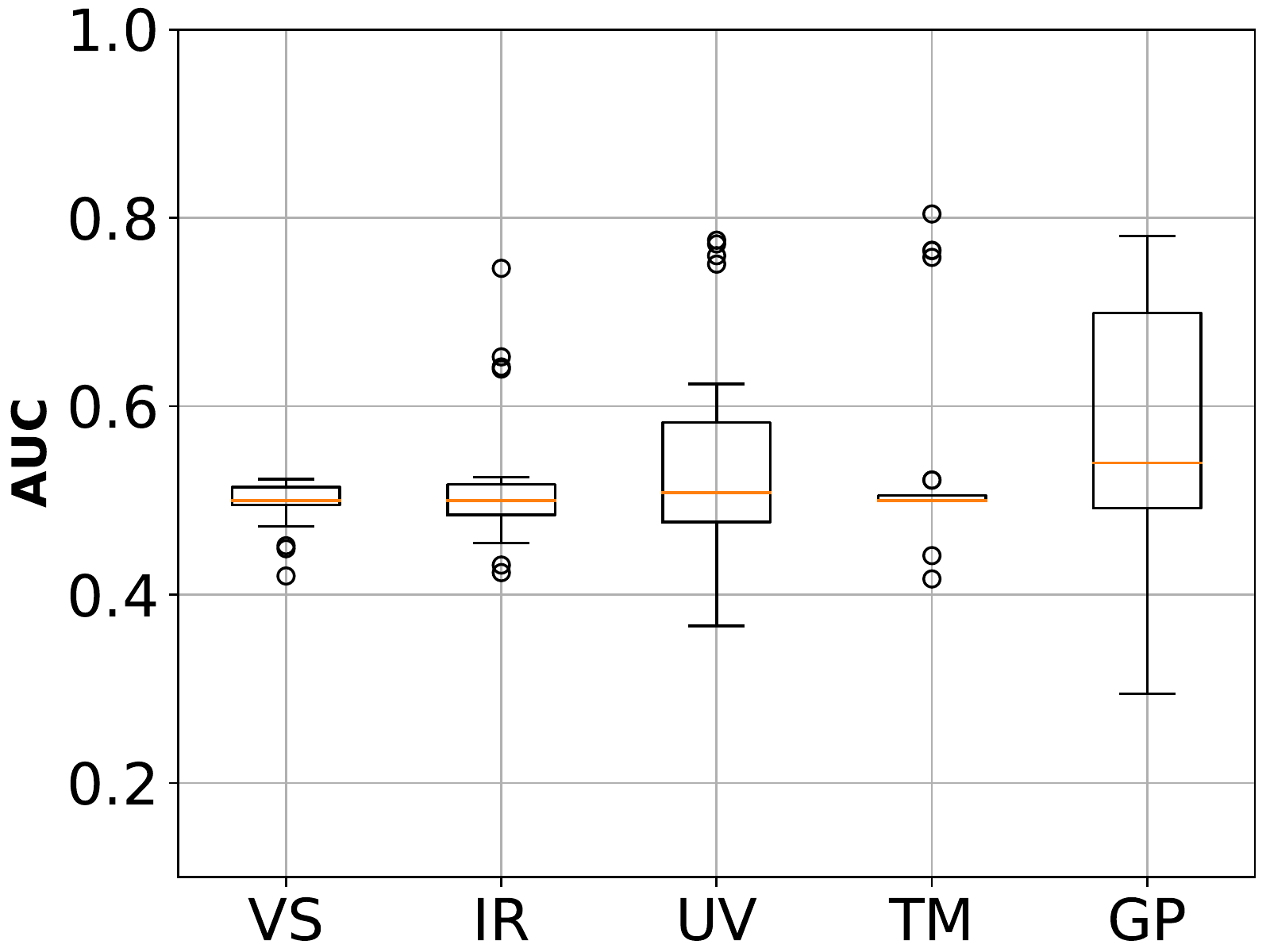}
		\caption{Case 1. Validation results}\label{fig:AVCC1}
			\vspace{0.1cm}
	\end{subfigure}
    \hfill
    \begin{subfigure}[b]{0.32\textwidth}
	\centering
		\includegraphics[width=1\textwidth,height=0.7\textwidth]{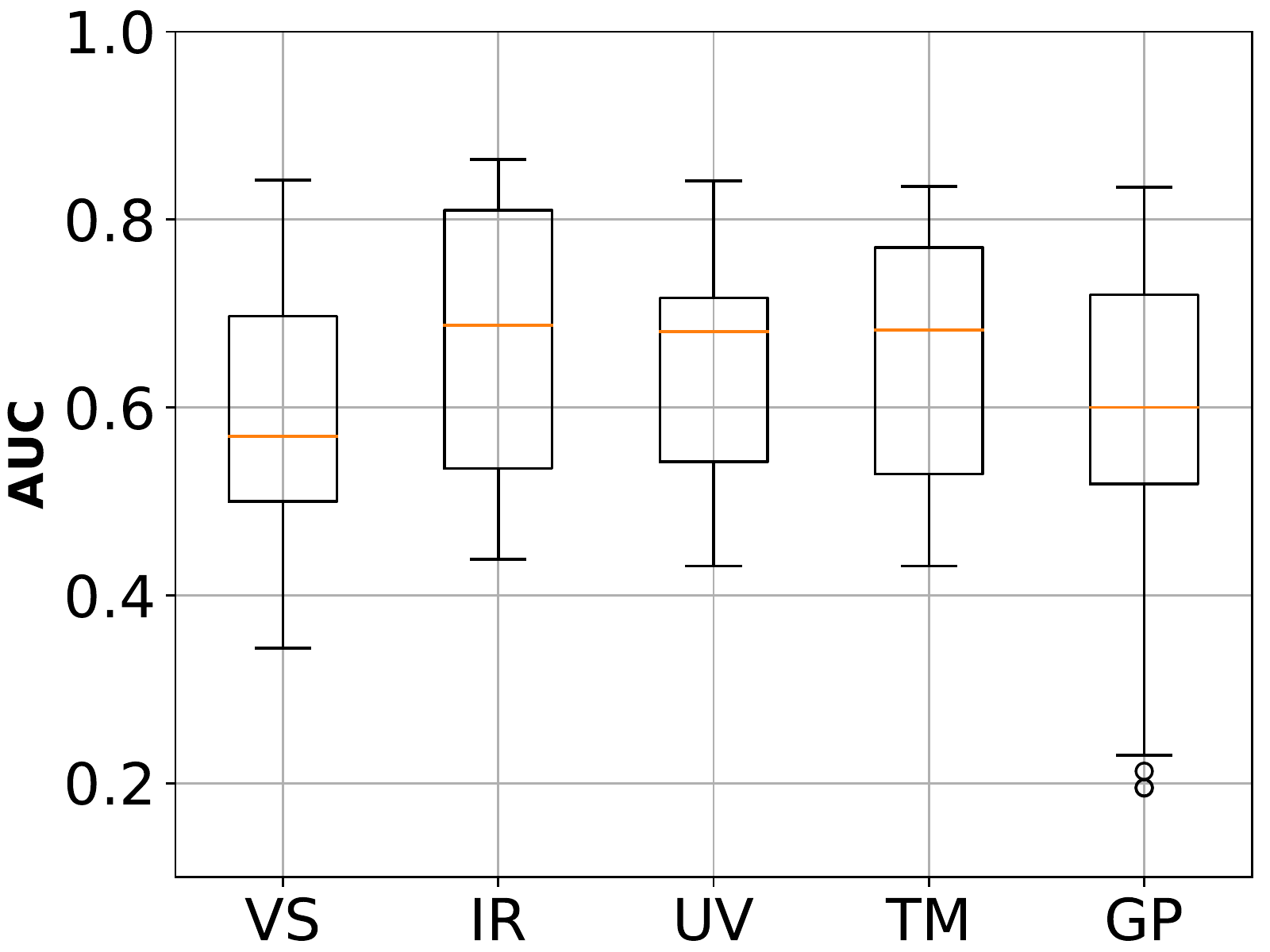}
		\caption{Case 2. Validation results}\label{fig:AVCC2}
			\vspace{0.1cm}
	\end{subfigure}
    \hfill
    \begin{subfigure}[b]{0.32\textwidth}
	\centering
		\includegraphics[width=1\textwidth,height=0.7\textwidth]{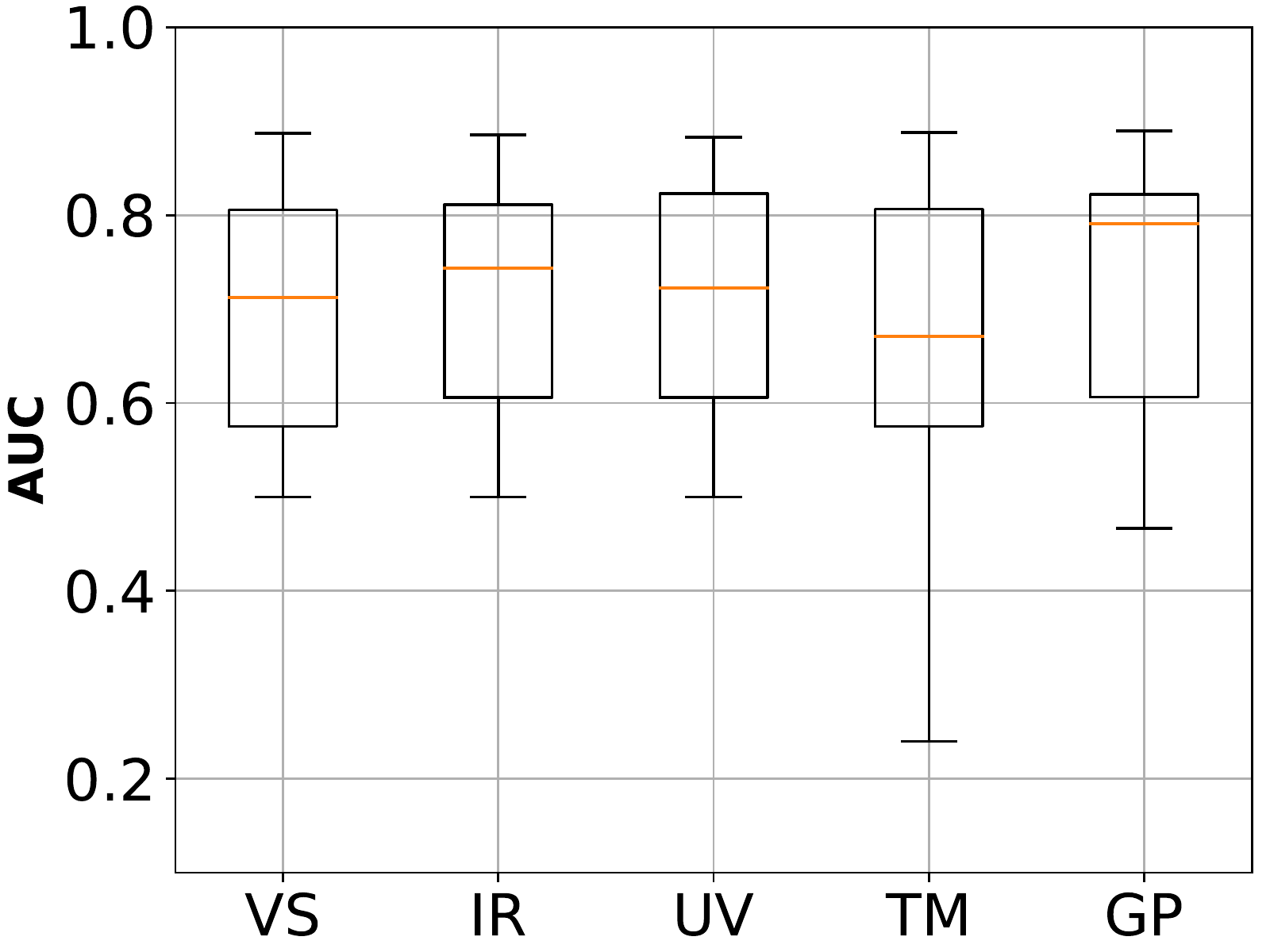}
		\caption{Case 3. Validation results}\label{fig:AVCC3}
			\vspace{0.1cm}
	\end{subfigure}
\caption{Distribution of the AUC for the cooperative decision models per type of sensor and per stage.}\label{fig:AUC_C}
\end{figure*}

A contrast between tables \ref{tab:Train} and \ref{tab:Validate} show us some behaviours. First, the best accuracy value among training and validation stages decrements for cases 1 and 2, in contrast, results in case 3 have a similar performance. Second, the best RMSE value in the validation results (see table \ref{tab:Validate}) belong to case 3, in both cases, local and cooperative model decisions. Although, the best performance for the RMSE metric in the training case belongs to case 2. And third, the highest AUC values per table, it means the best results of the AUC metric, are related to case 3. This last observation is also discernible in figure \ref{fig:AUC_C}, where the mean values (orange line) of box-plot of the AUC results for case 3, are higher than the mean values from cases 1 and 2.

\begin{figure*}[ht!]
\centering
	\begin{subfigure}[b]{0.32\textwidth}
	\centering
		\includegraphics[width=1\textwidth,height=0.62\textwidth]{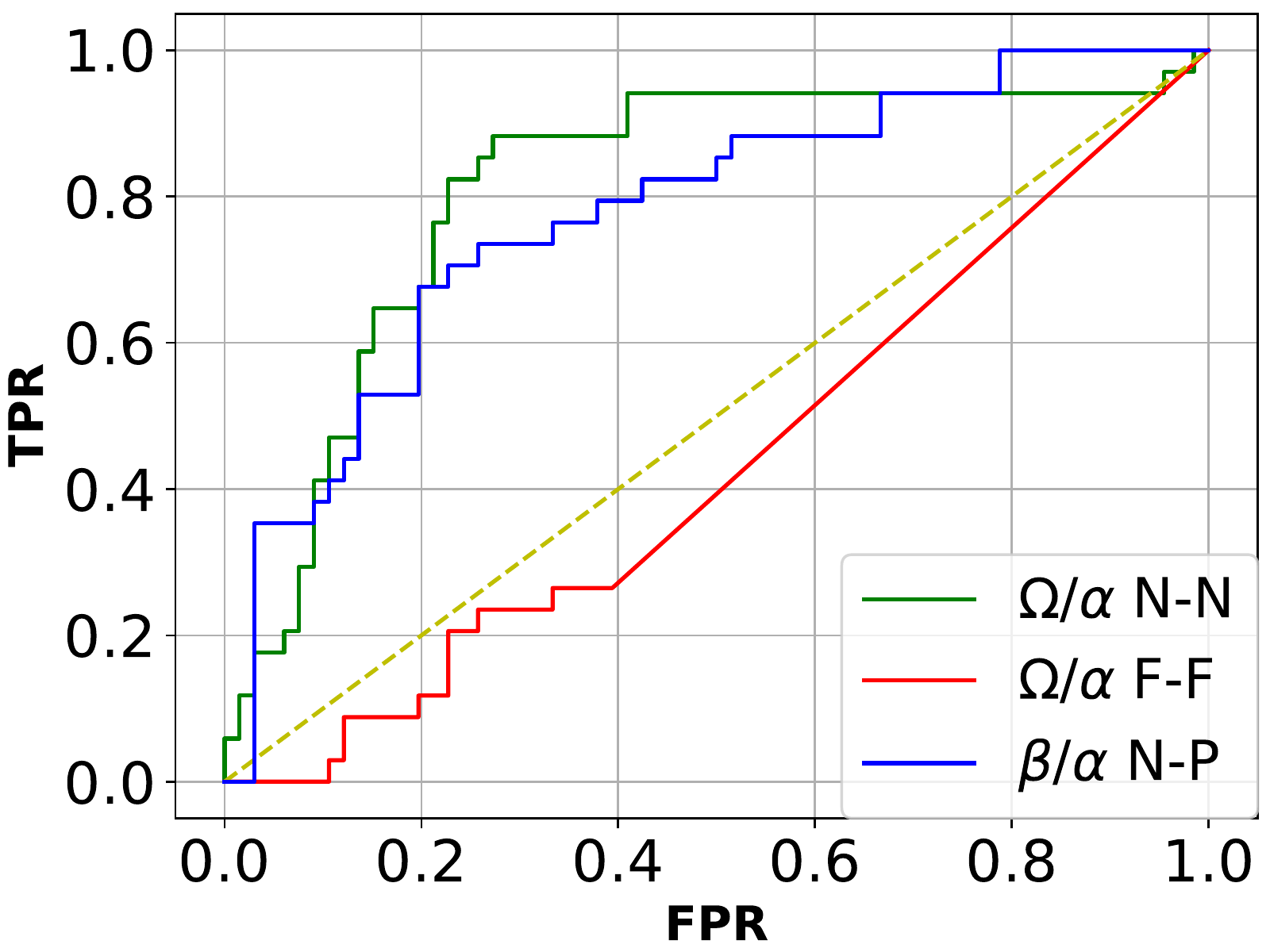}
		\caption{VS -- case 1}\label{fig:TSVL1}
			\vspace{0.1cm}
	\end{subfigure}
    \hfill
    \hfill\begin{subfigure}[b]{0.32\textwidth}
	\centering
		\includegraphics[width=1\textwidth,height=0.62\textwidth]{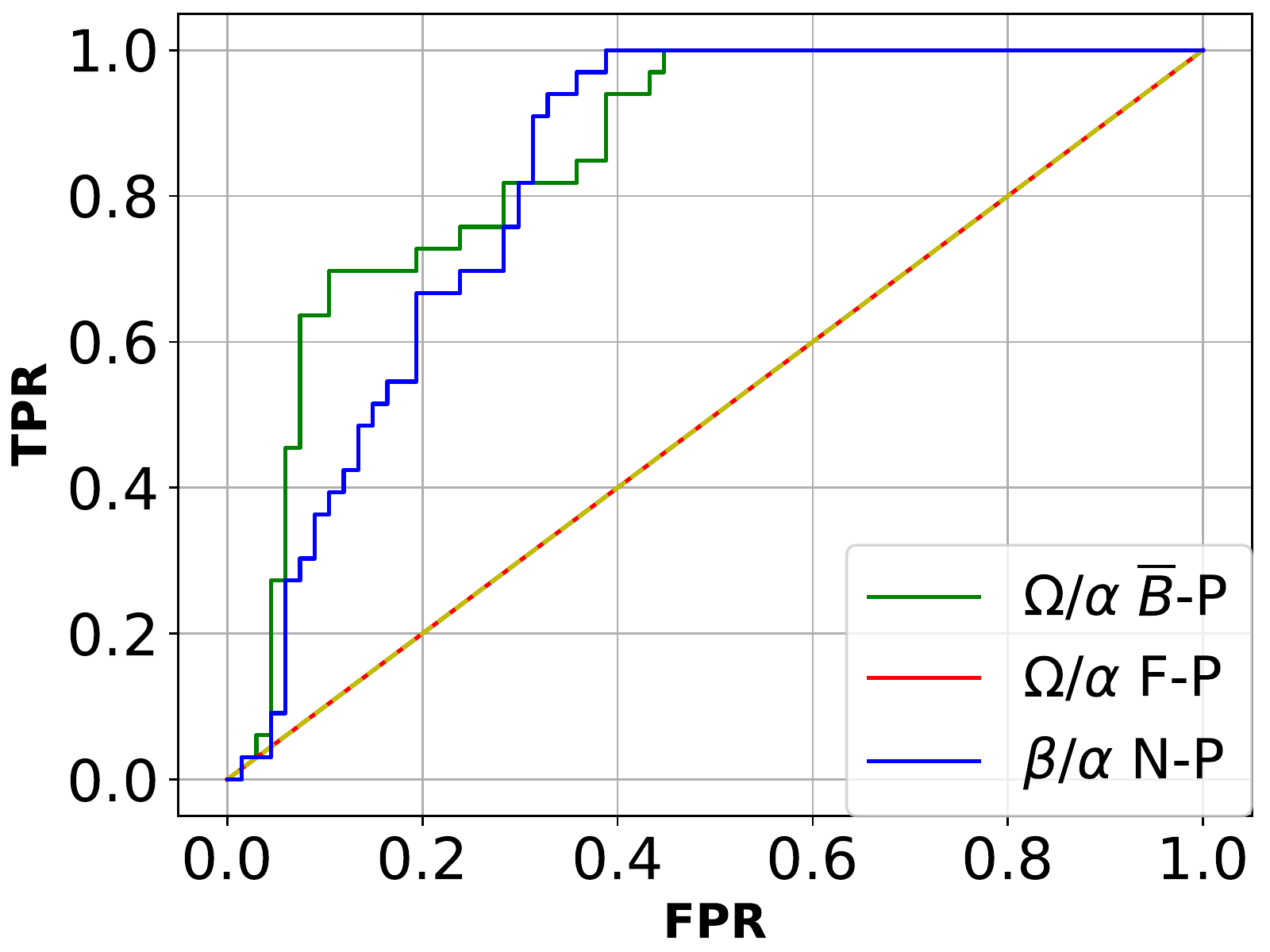}
		\caption{VS -- case 2}\label{fig:TSVL2}
			\vspace{0.1cm}
	\end{subfigure}
    \hfill
    \begin{subfigure}[b]{0.32\textwidth}
	\centering
		\includegraphics[width=1\textwidth,height=0.62\textwidth]{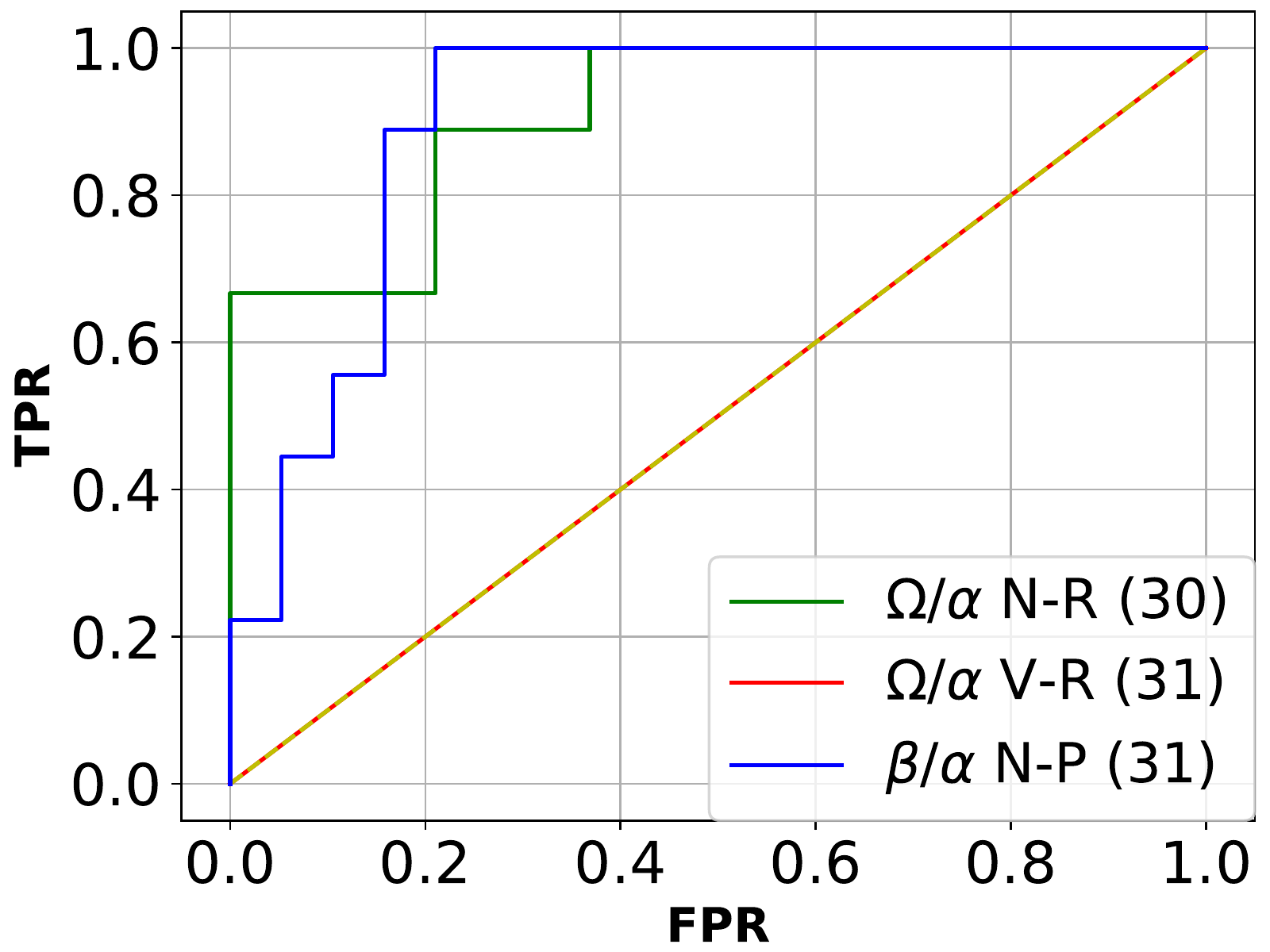}
		\caption{VS -- case 3}\label{fig:TSVL3}
			\vspace{0.1cm}
	\end{subfigure}
    \hfill
    \begin{subfigure}[b]{0.32\textwidth}
	\centering
		\includegraphics[width=1\textwidth,height=0.62\textwidth]{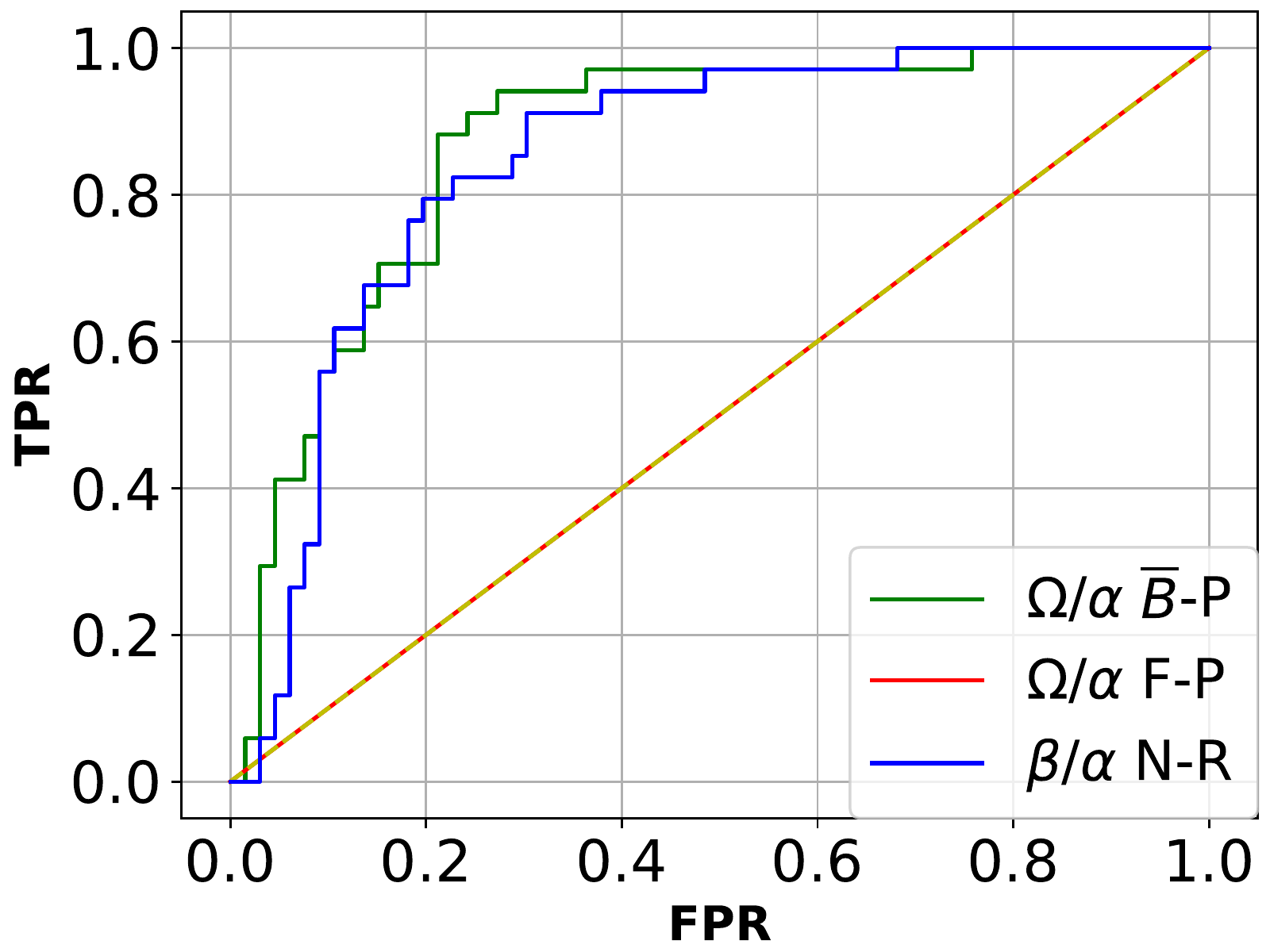}
		\caption{IR -- case 1}\label{fig:TIRL1}
			\vspace{0.1cm}
	\end{subfigure}
    \hfill
    \begin{subfigure}[b]{0.32\textwidth}
	\centering
		\includegraphics[width=1\textwidth,height=0.62\textwidth]{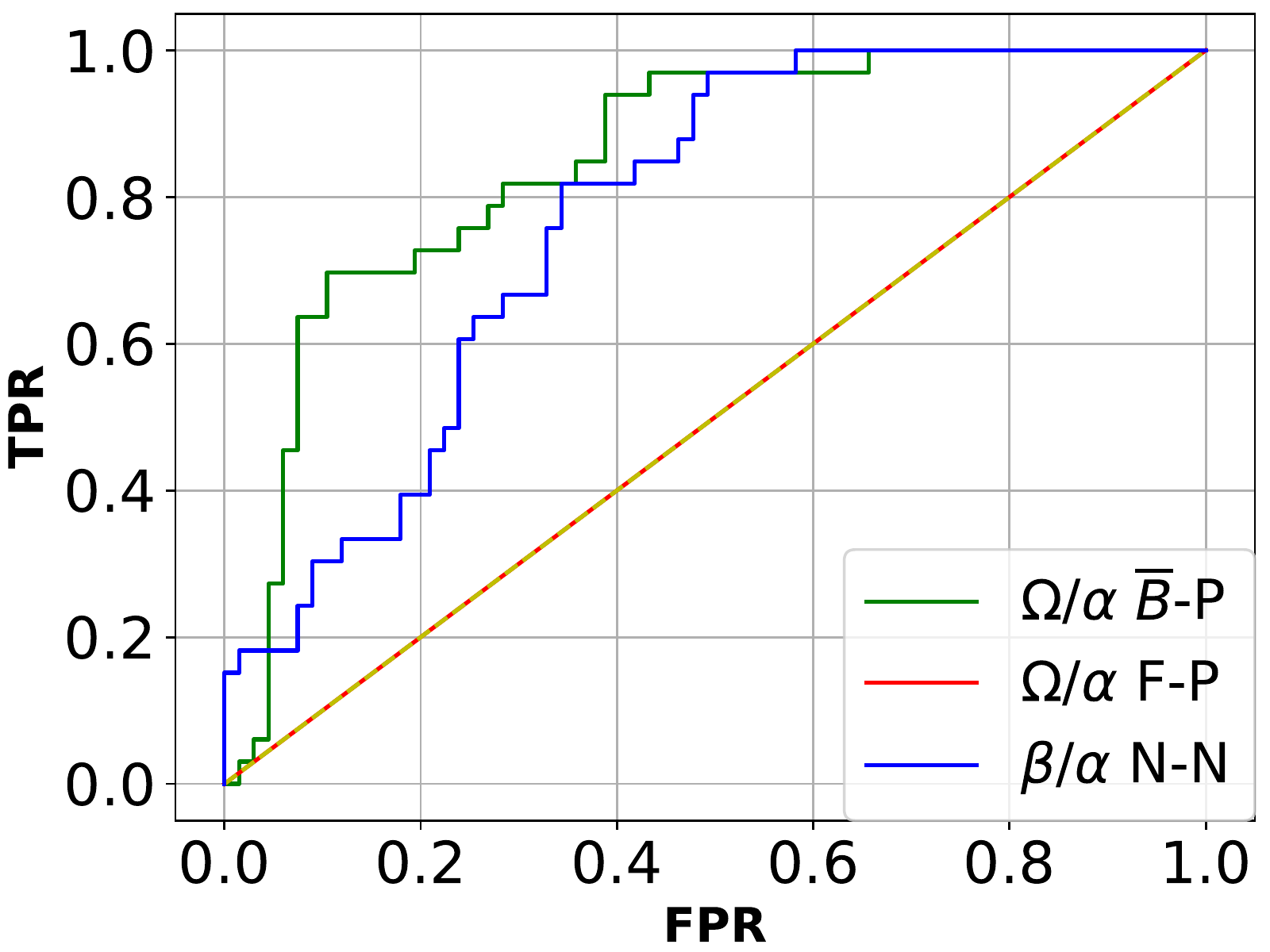}
		\caption{IR -- case 2}\label{fig:TIRL2}
			\vspace{0.1cm}
	\end{subfigure}
    \hfill
    \begin{subfigure}[b]{0.32\textwidth}
	\centering
		\includegraphics[width=1\textwidth,height=0.62\textwidth]{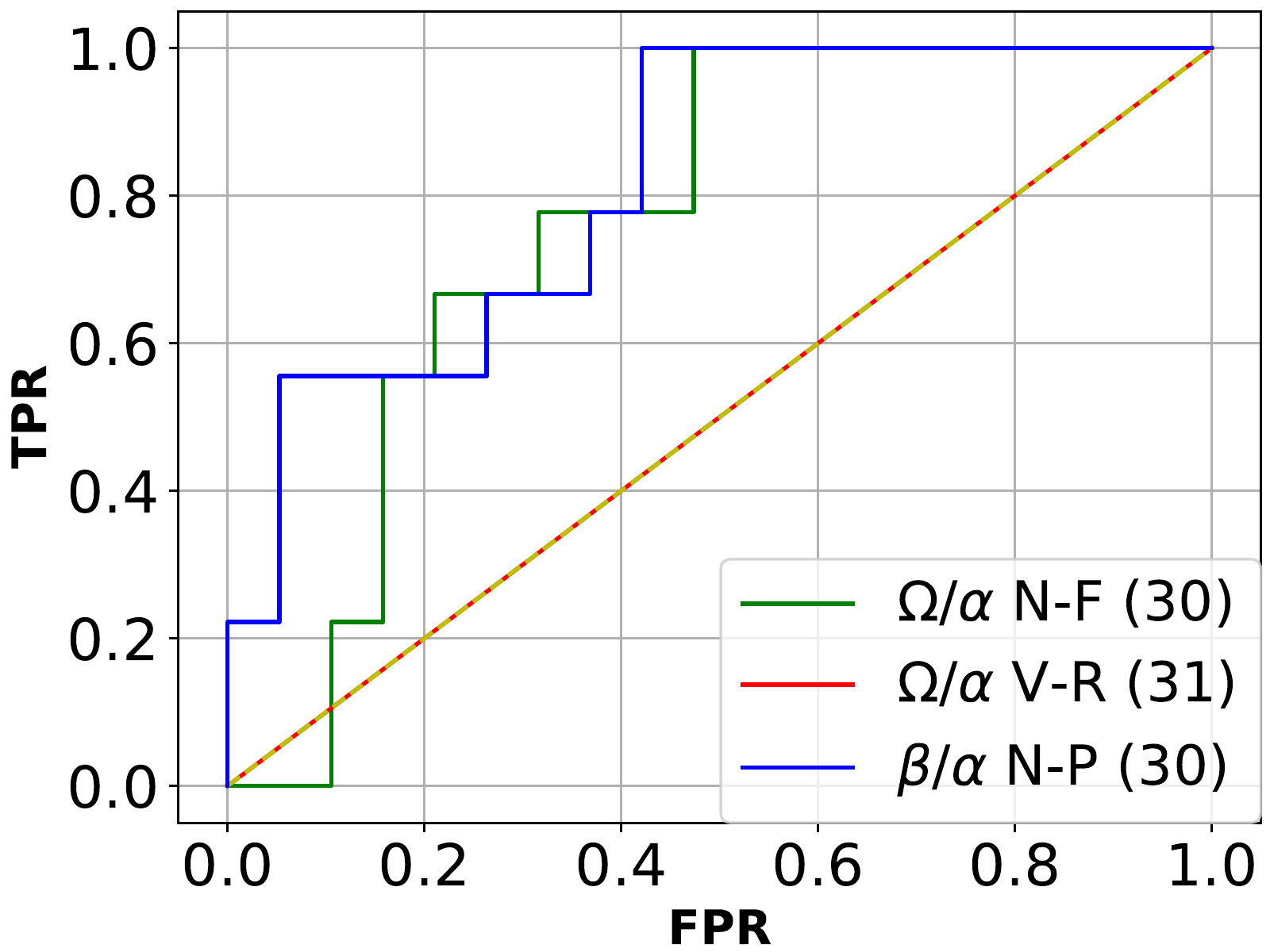}
		\caption{IR -- case 3}\label{fig:TIRL3}
			\vspace{0.1cm}
	\end{subfigure}
    \hfill
    \begin{subfigure}[b]{0.32\textwidth}
	\centering
		\includegraphics[width=1\textwidth,height=0.62\textwidth]{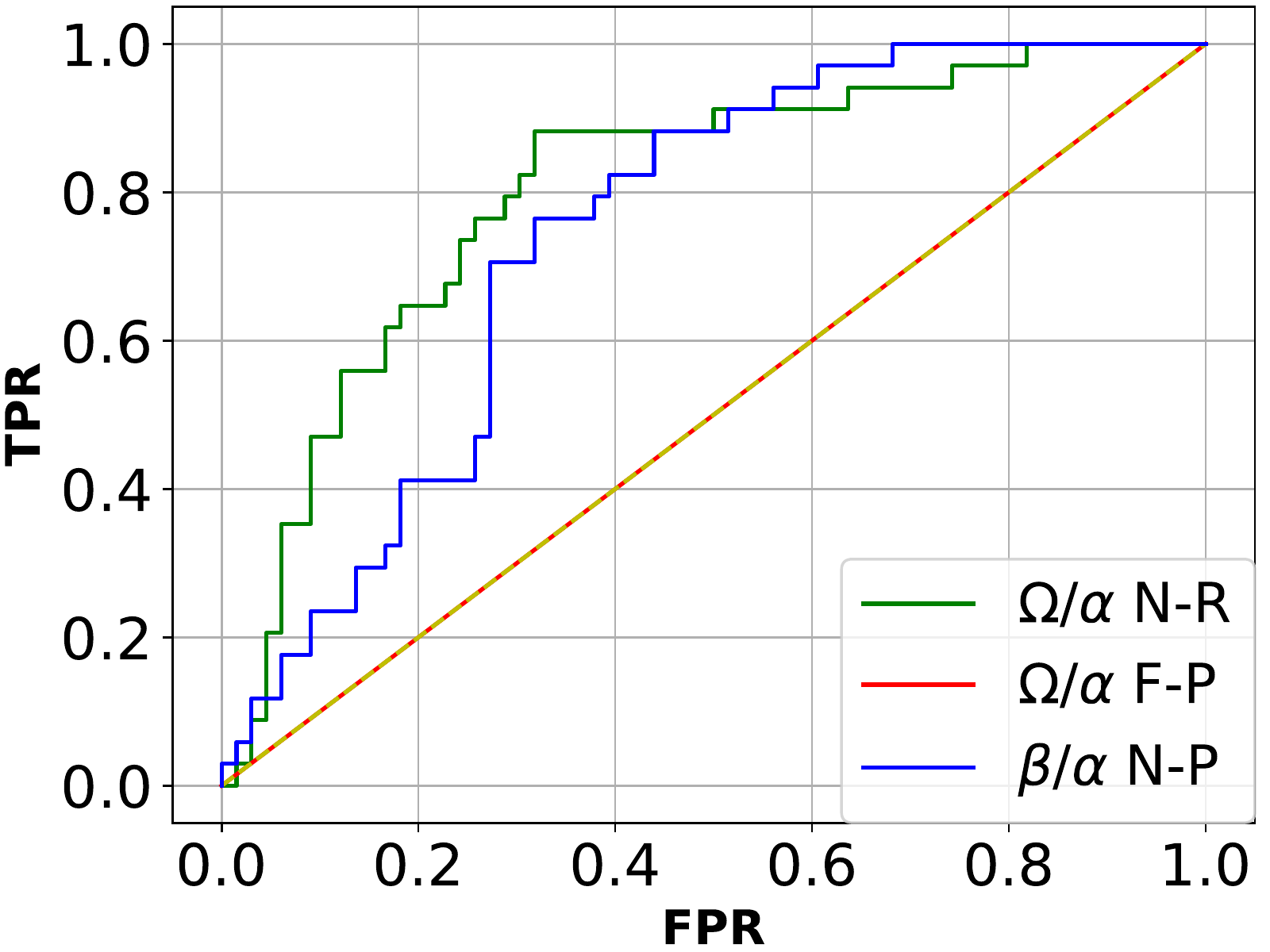}
		\caption{UV -- case 1}\label{fig:TUVL1}
			\vspace{0.1cm}
	\end{subfigure}
    \hfill
    \begin{subfigure}[b]{0.32\textwidth}
	\centering
		\includegraphics[width=1\textwidth,height=0.62\textwidth]{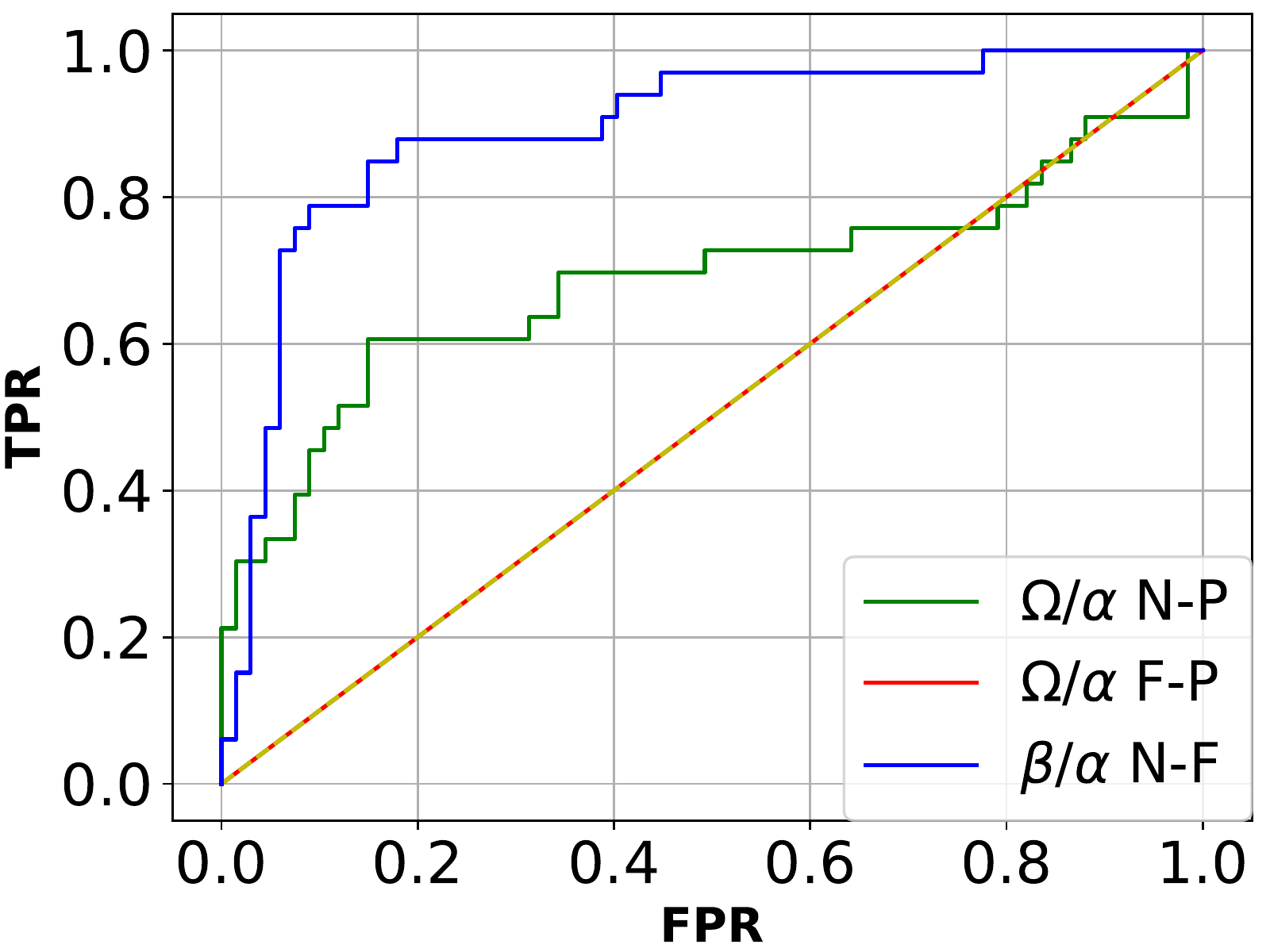}
		\caption{UV -- case 2}\label{fig:TUVL2}
			\vspace{0.1cm}
	\end{subfigure}
    \hfill
    \begin{subfigure}[b]{0.32\textwidth}
	\centering
		\includegraphics[width=1\textwidth,height=0.62\textwidth]{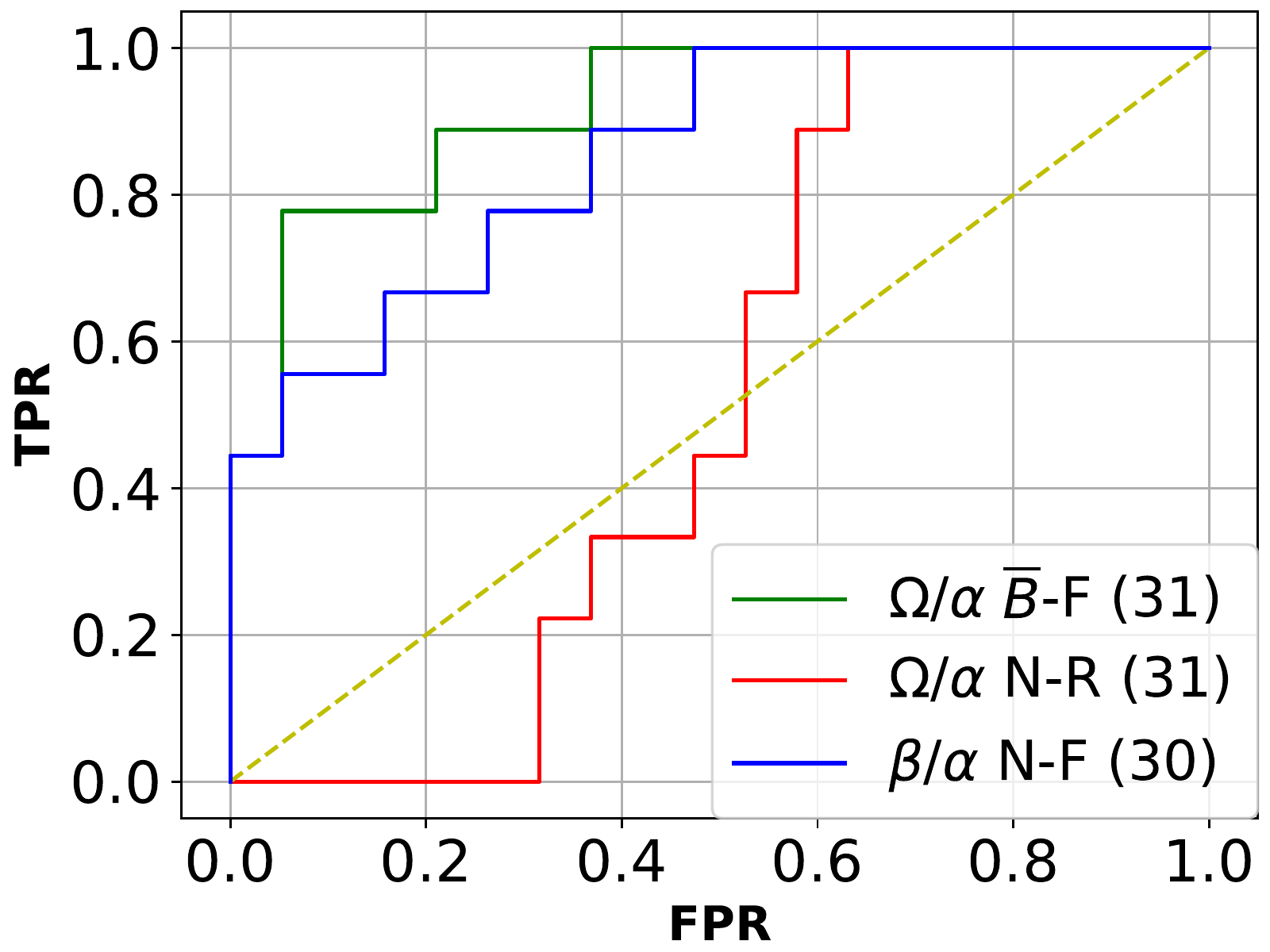}
		\caption{UV -- case 3}\label{fig:TUVL3}
			\vspace{0.1cm}
	\end{subfigure}
    \hfill
    \begin{subfigure}[b]{0.32\textwidth}
	\centering
		\includegraphics[width=1\textwidth,height=0.62\textwidth]{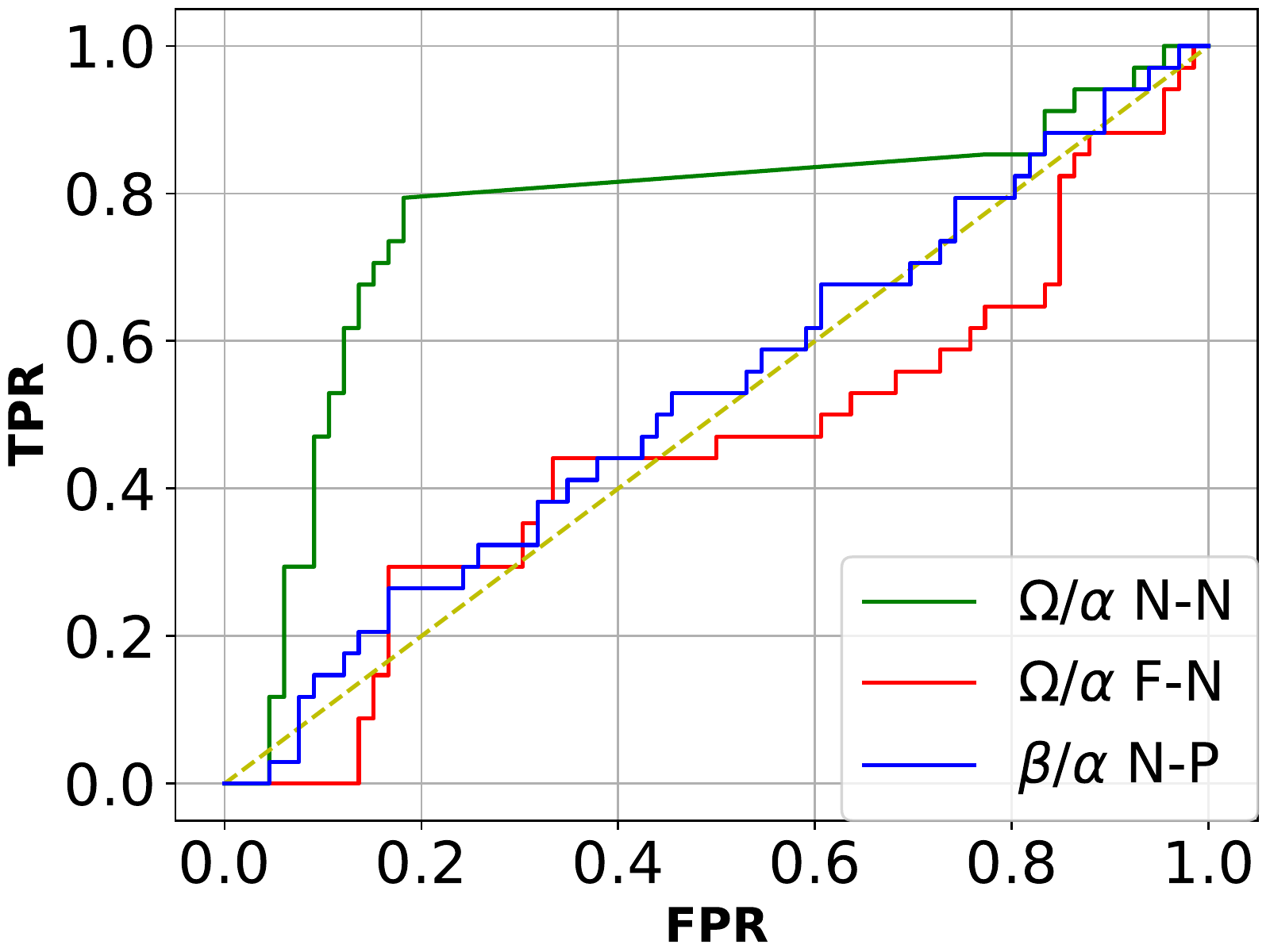}
		\caption{TM -- case 1}\label{fig:TTML1}
			\vspace{0.1cm}
	\end{subfigure}
    \hfill
    \begin{subfigure}[b]{0.32\textwidth}
	\centering
		\includegraphics[width=1\textwidth,height=0.62\textwidth]{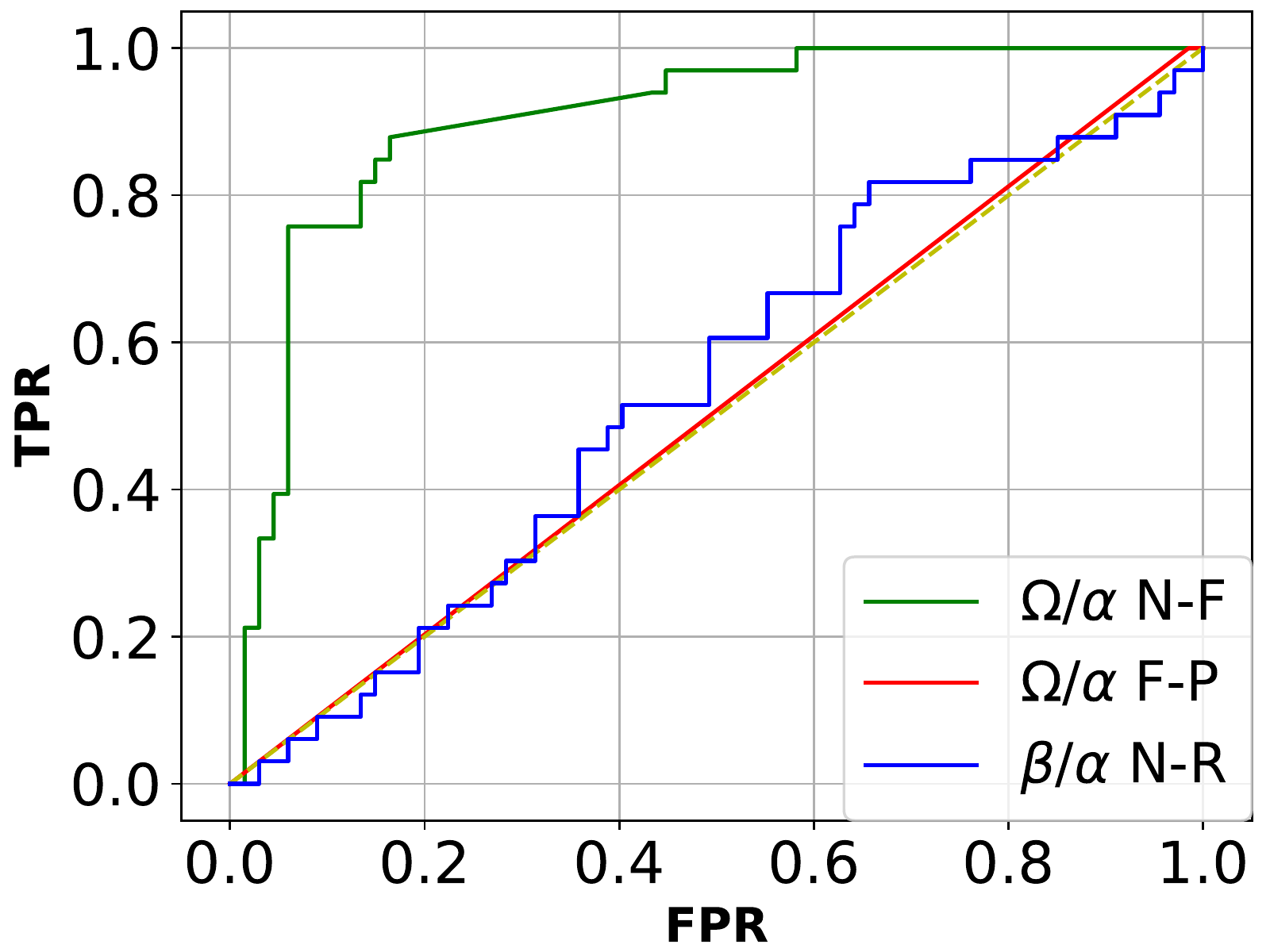}
		\caption{TM -- case 2}\label{fig:TTML2}
			\vspace{0.1cm}
	\end{subfigure}
    \hfill
    \begin{subfigure}[b]{0.32\textwidth}
	\centering
		\includegraphics[width=1\textwidth,height=0.62\textwidth]{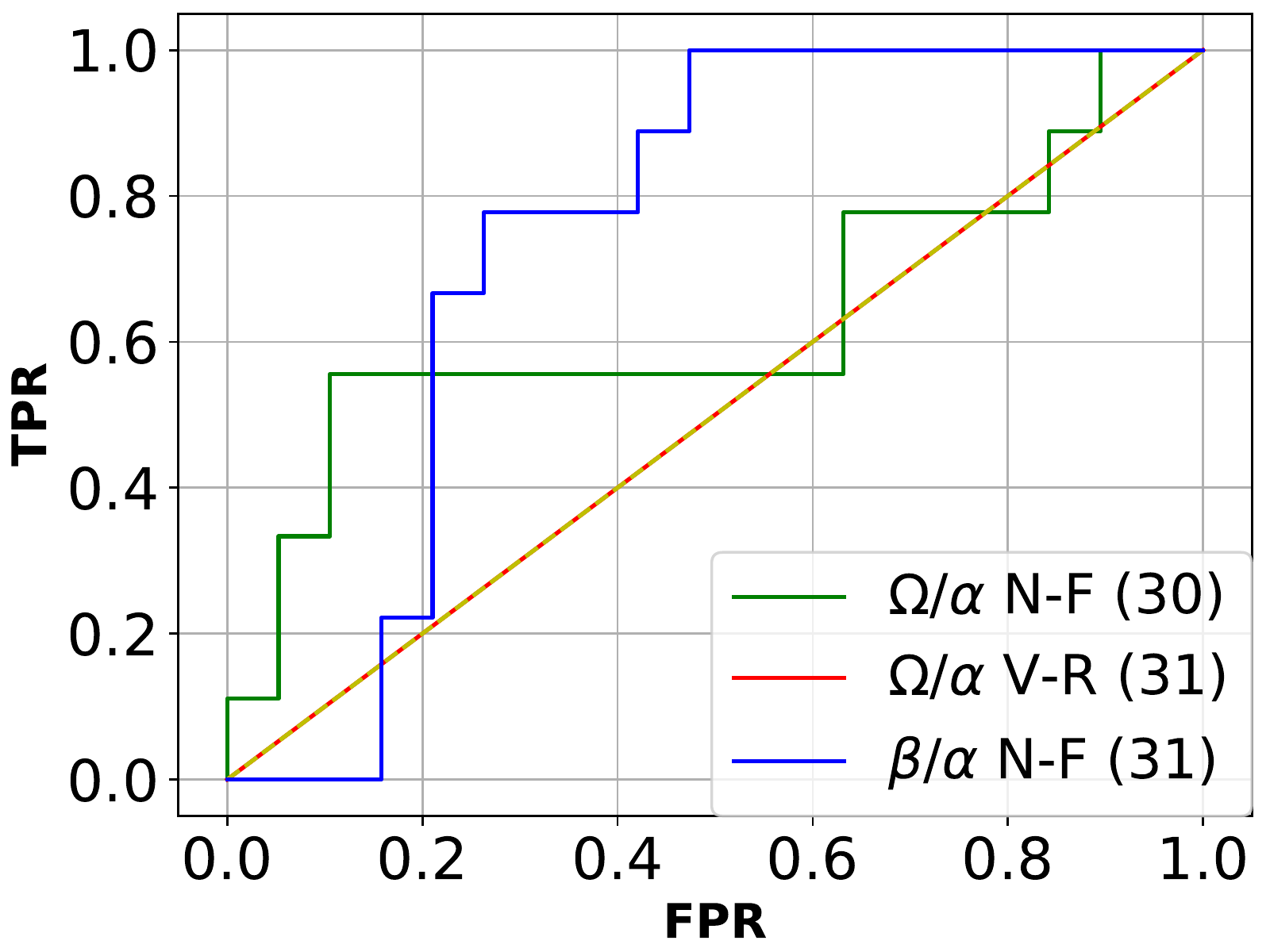}
		\caption{TM -- case 3}\label{fig:TTML3}
			\vspace{0.1cm}
	\end{subfigure}
    \hfill
    \begin{subfigure}[b]{0.32\textwidth}
	\centering
		\includegraphics[width=1\textwidth,height=0.62\textwidth]{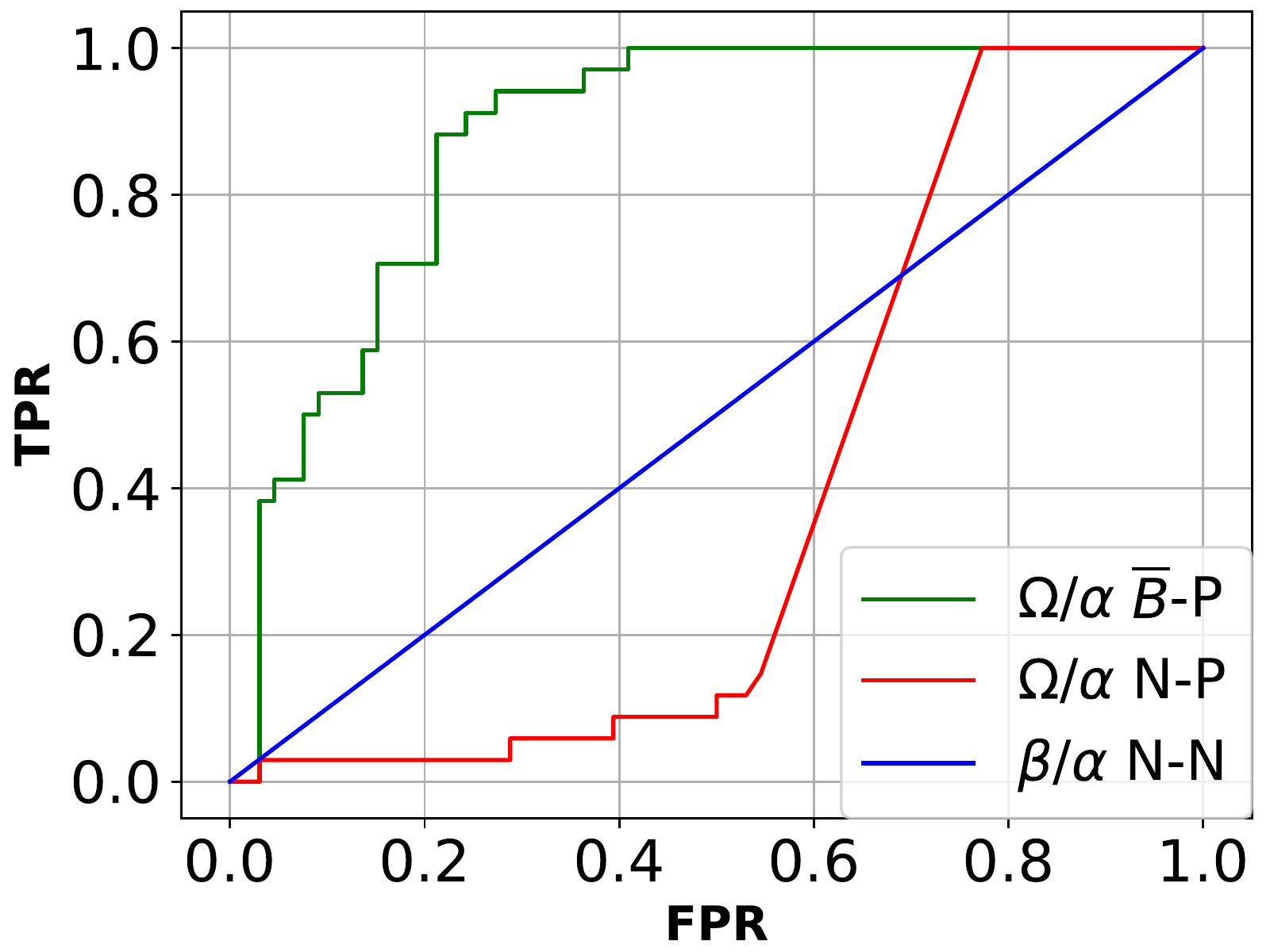}
		\caption{GP -- case 1}\label{fig:TGPL1}
			\vspace{0.1cm}
	\end{subfigure}
    \hfill
    \begin{subfigure}[b]{0.32\textwidth}
	\centering
		\includegraphics[width=1\textwidth,height=0.62\textwidth]{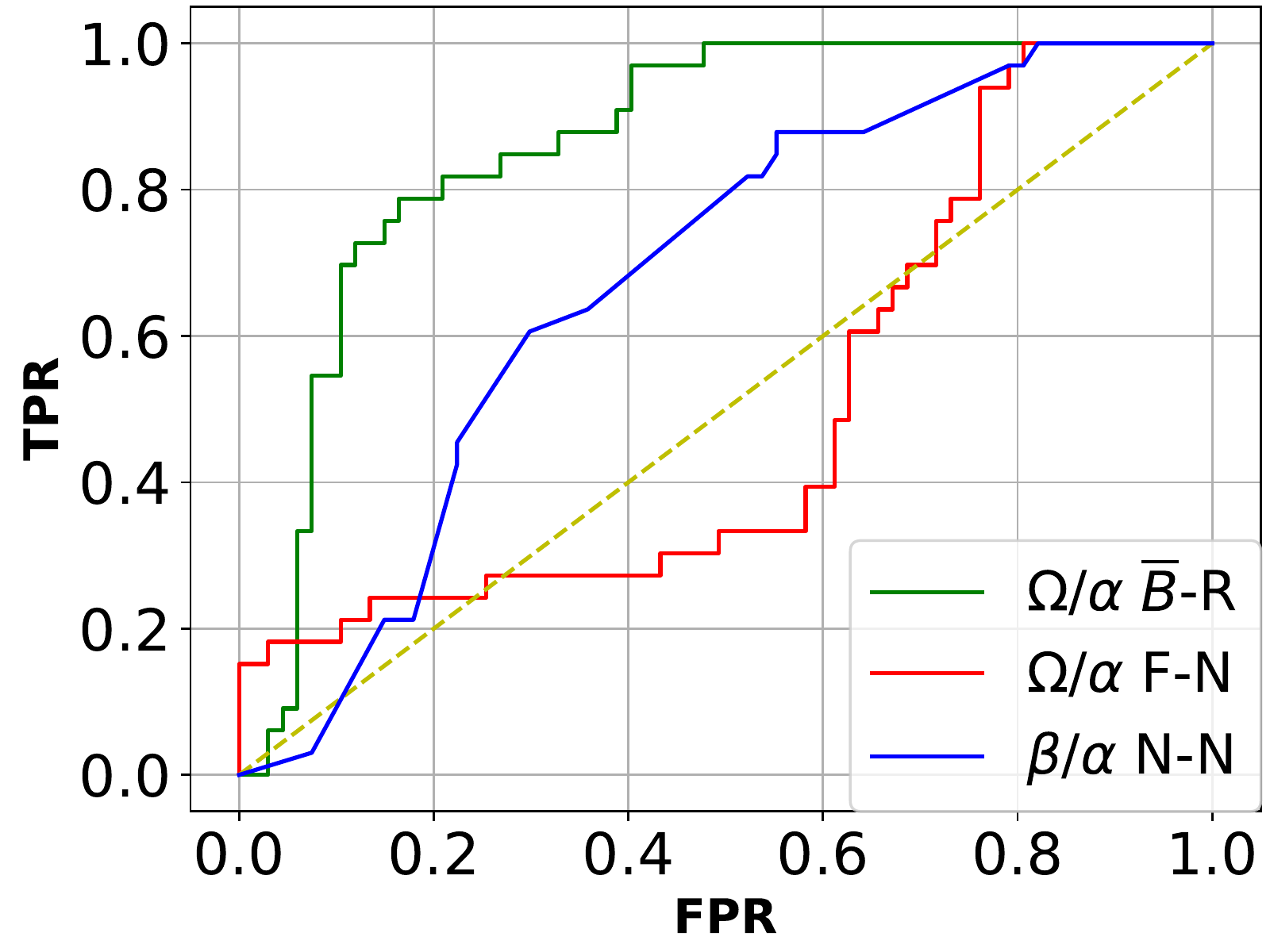}
		\caption{GP -- case 2}\label{fig:TGPL2}
			\vspace{0.1cm}
	\end{subfigure}
    \hfill
    \begin{subfigure}[b]{0.32\textwidth}
	\centering
		\includegraphics[width=1\textwidth,height=0.62\textwidth]{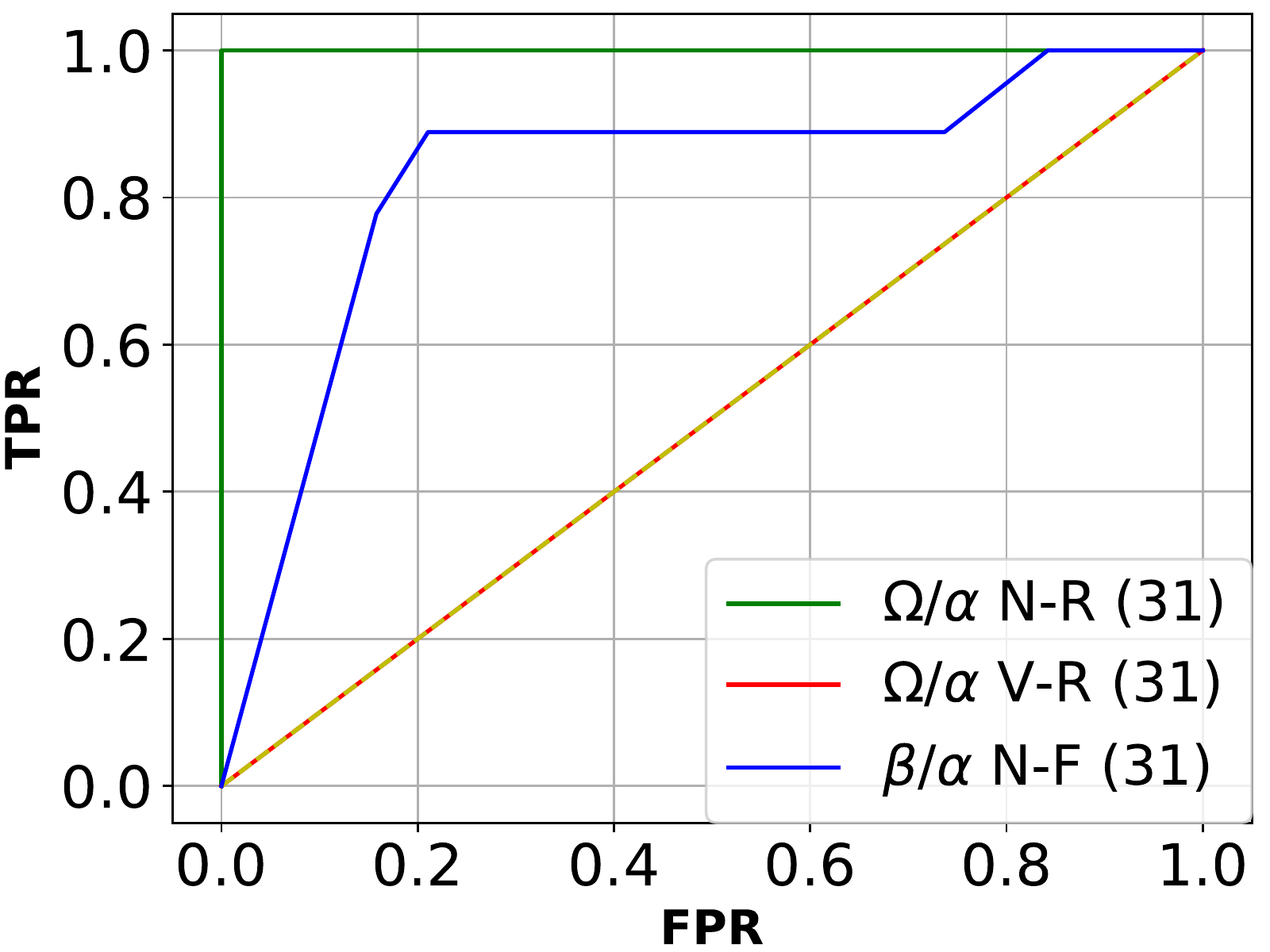}
		\caption{GP -- case 3}\label{fig:TGPL3}
			\vspace{0.1cm}
	\end{subfigure}
\caption{Selected ROC curves from training data set per sensor and per case. Green curve belongs to the best ACC result of the cooperative decision-making, the red one to the worst cooperative result, and the blue one to the best local decision-making result.}\label{fig:ROC_T}
\end{figure*}

\begin{figure*}[ht!]
\centering
	\begin{subfigure}[b]{0.32\textwidth}
	\centering
		\includegraphics[width=1\textwidth,height=0.62\textwidth]{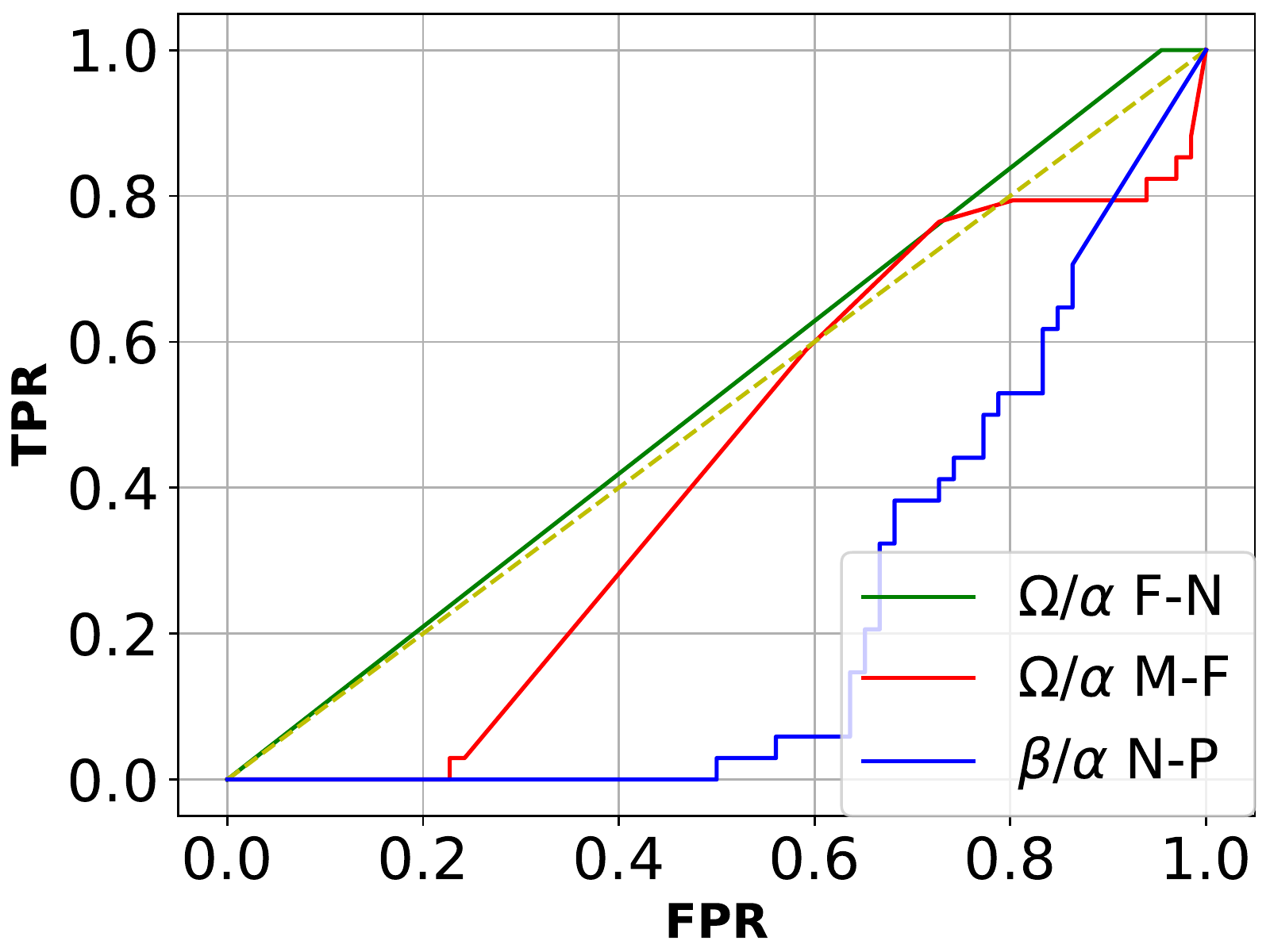}
		\caption{VS -- case 1}\label{fig:VSVL1}
			\vspace{0.1cm}
	\end{subfigure}
    \hfill
    \hfill\begin{subfigure}[b]{0.32\textwidth}
	\centering
		\includegraphics[width=1\textwidth,height=0.62\textwidth]{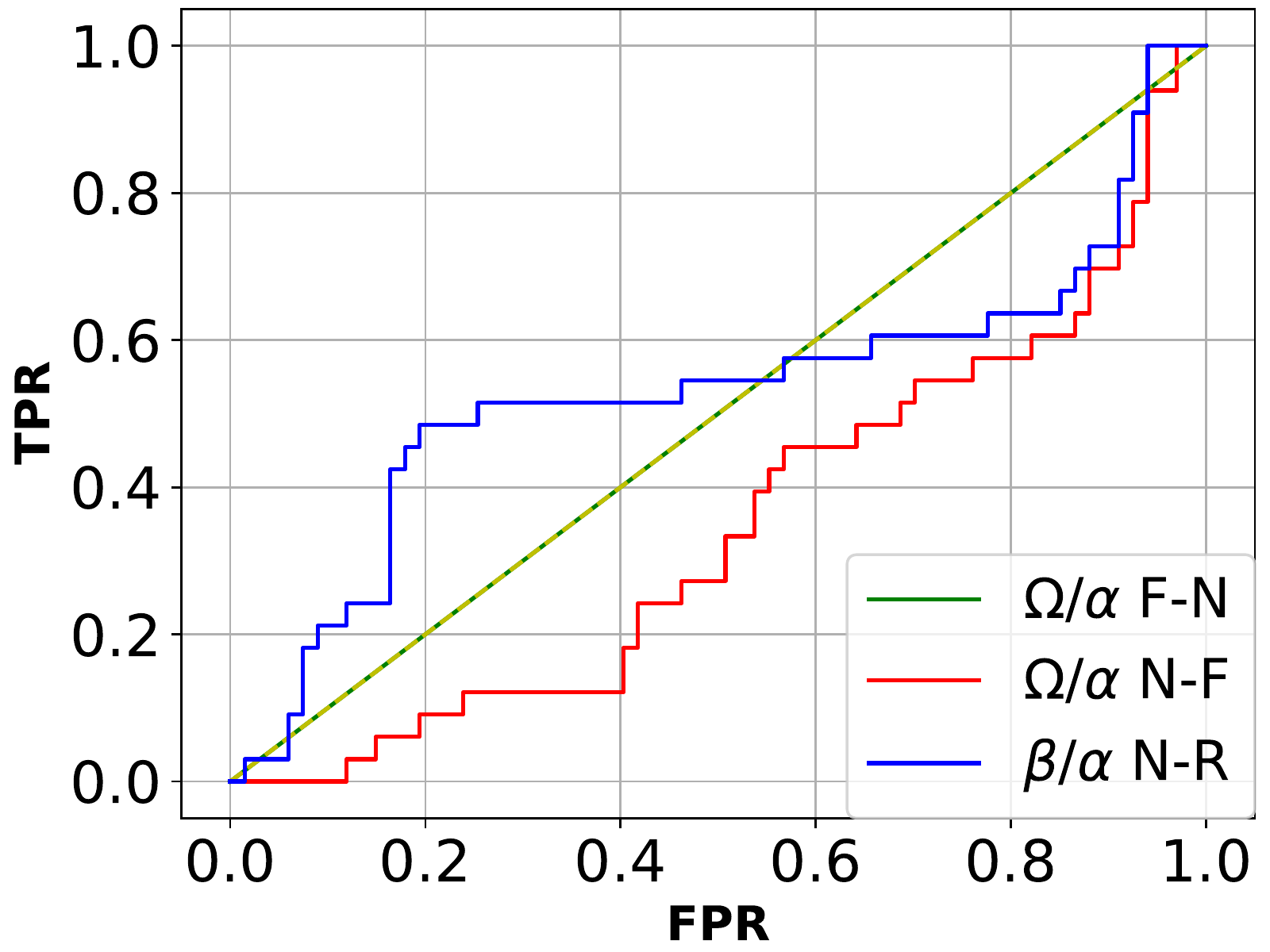}
		\caption{VS -- case 2}\label{fig:VSVL2}
			\vspace{0.1cm}
	\end{subfigure}
    \hfill
    \begin{subfigure}[b]{0.32\textwidth}
	\centering
		\includegraphics[width=1\textwidth,height=0.62\textwidth]{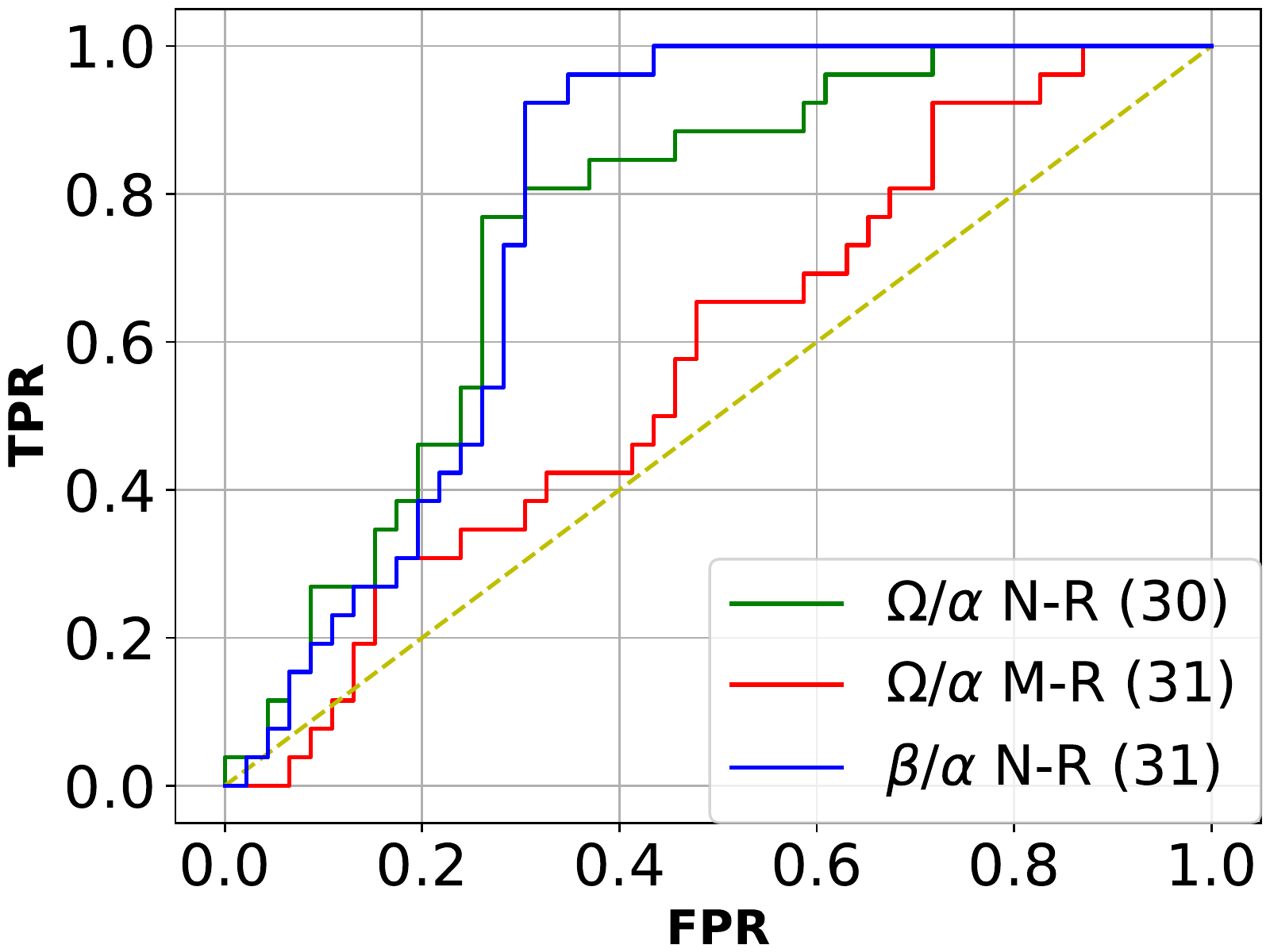}
		\caption{VS -- case 3}\label{fig:VSVL3}
			\vspace{0.1cm}
	\end{subfigure}
    \hfill
    \begin{subfigure}[b]{0.32\textwidth}
	\centering
		\includegraphics[width=1\textwidth,height=0.62\textwidth]{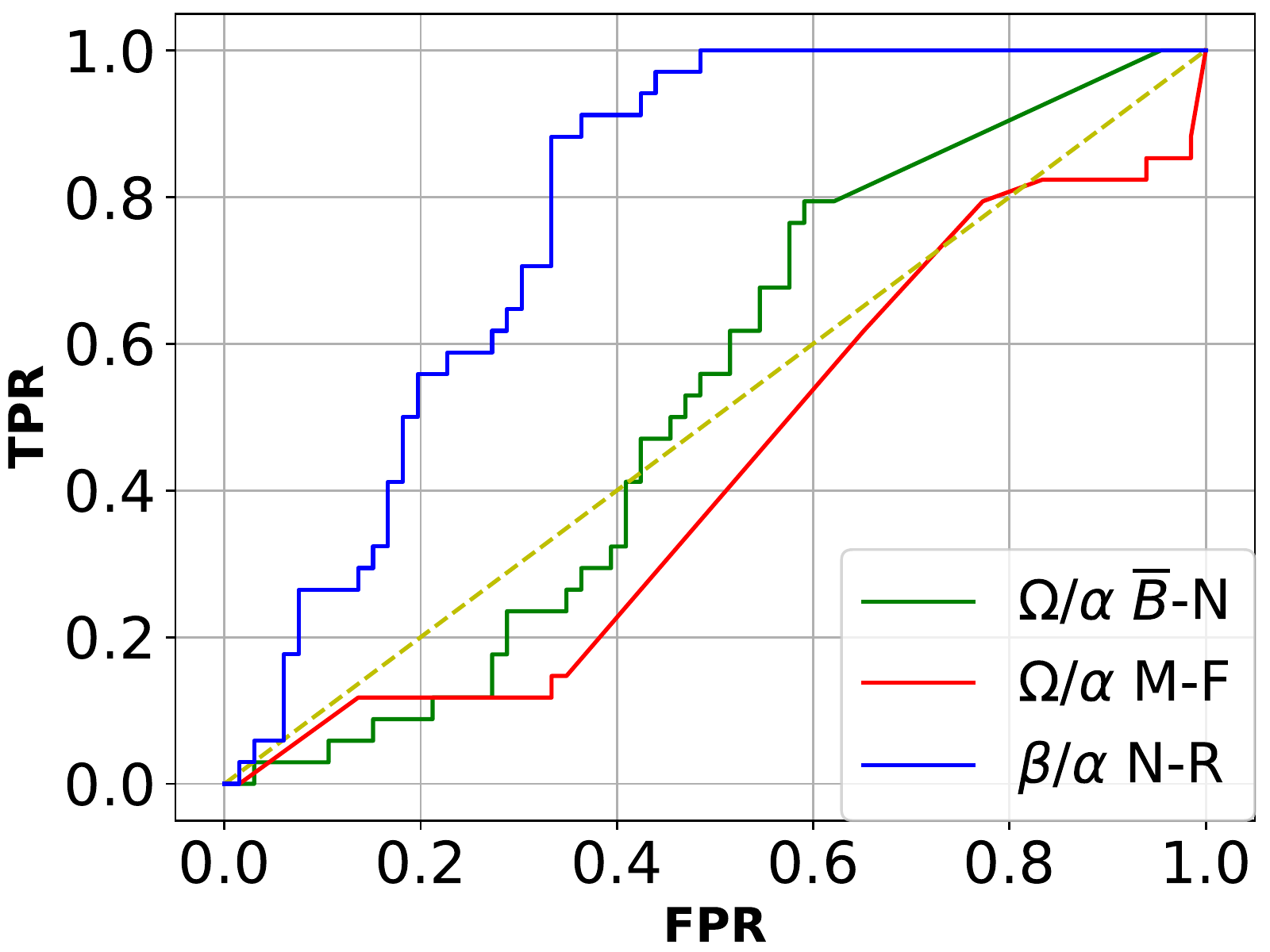}
		\caption{IR -- case 1}\label{fig:VIRL1}
			\vspace{0.1cm}
	\end{subfigure}
    \hfill
    \begin{subfigure}[b]{0.32\textwidth}
	\centering
		\includegraphics[width=1\textwidth,height=0.62\textwidth]{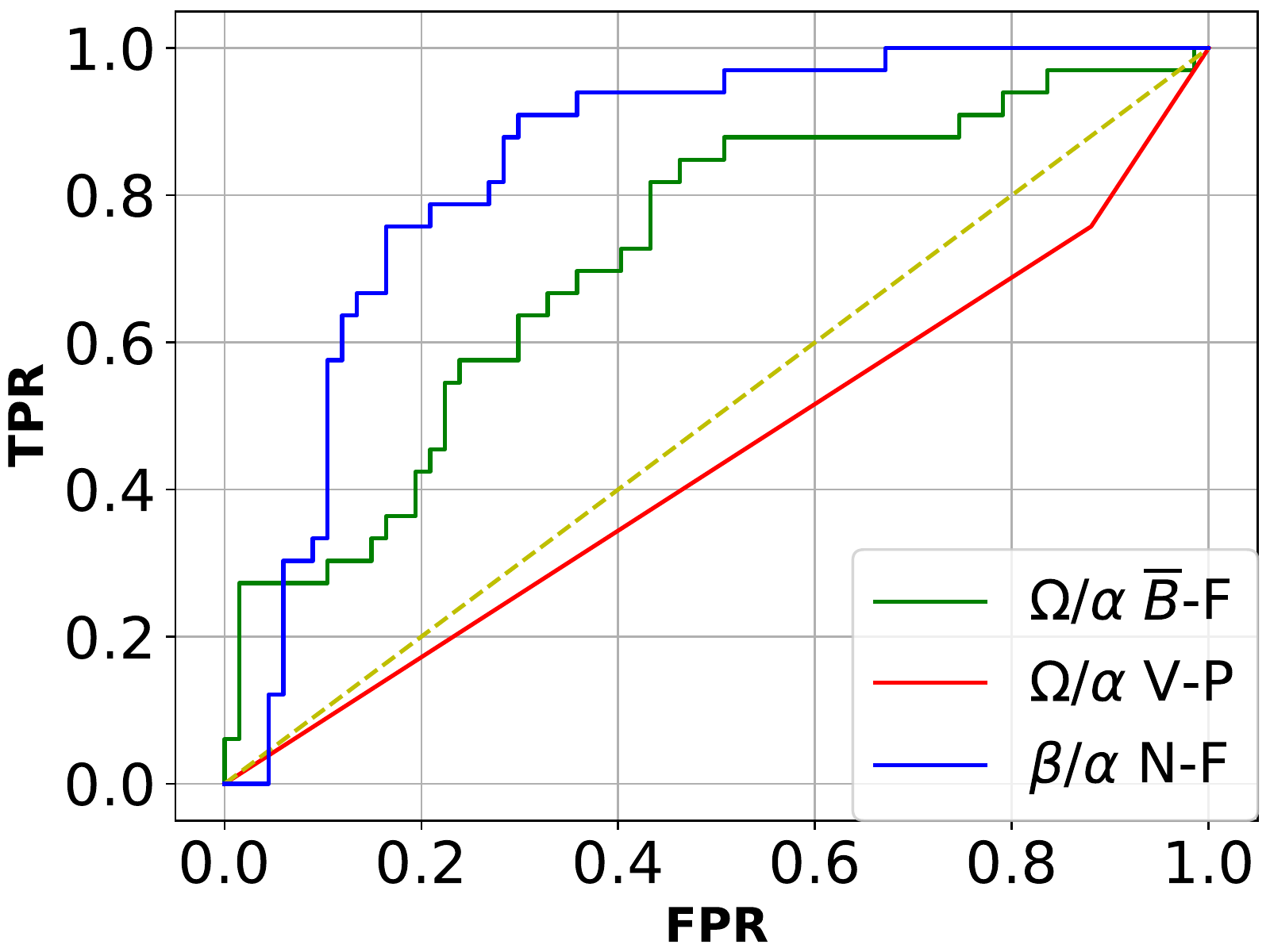}
		\caption{IR -- case 2}\label{fig:VIRL2}
			\vspace{0.1cm}
	\end{subfigure}
    \hfill
    \begin{subfigure}[b]{0.32\textwidth}
	\centering
		\includegraphics[width=1\textwidth,height=0.62\textwidth]{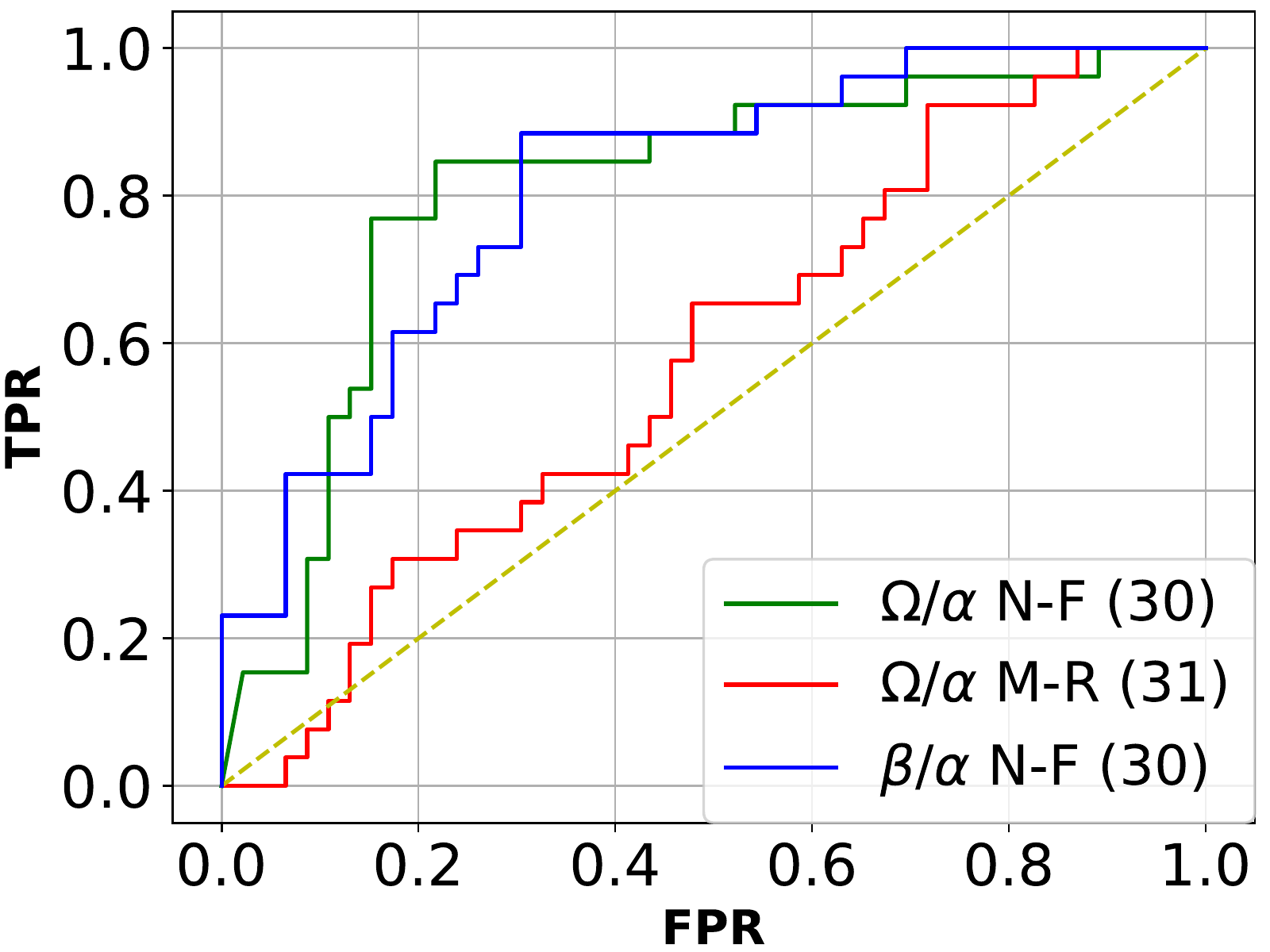}
		\caption{IR -- case 3}\label{fig:VIRL3}
			\vspace{0.1cm}
	\end{subfigure}
    \hfill
    \begin{subfigure}[b]{0.32\textwidth}
	\centering
		\includegraphics[width=1\textwidth,height=0.62\textwidth]{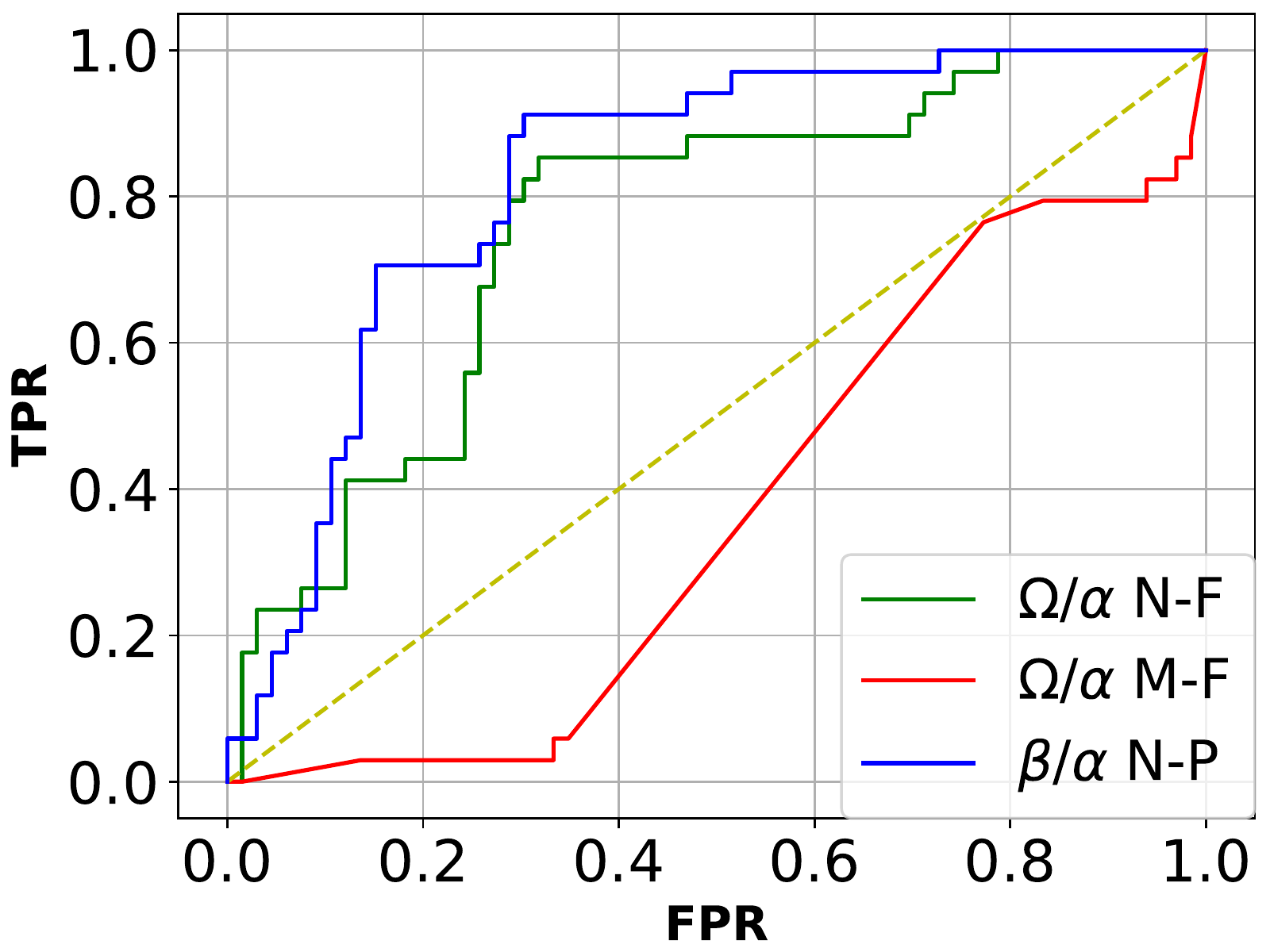}
		\caption{UV -- case 1}\label{fig:VUVL1}
			\vspace{0.1cm}
	\end{subfigure}
    \hfill
    \begin{subfigure}[b]{0.32\textwidth}
	\centering
		\includegraphics[width=1\textwidth,height=0.62\textwidth]{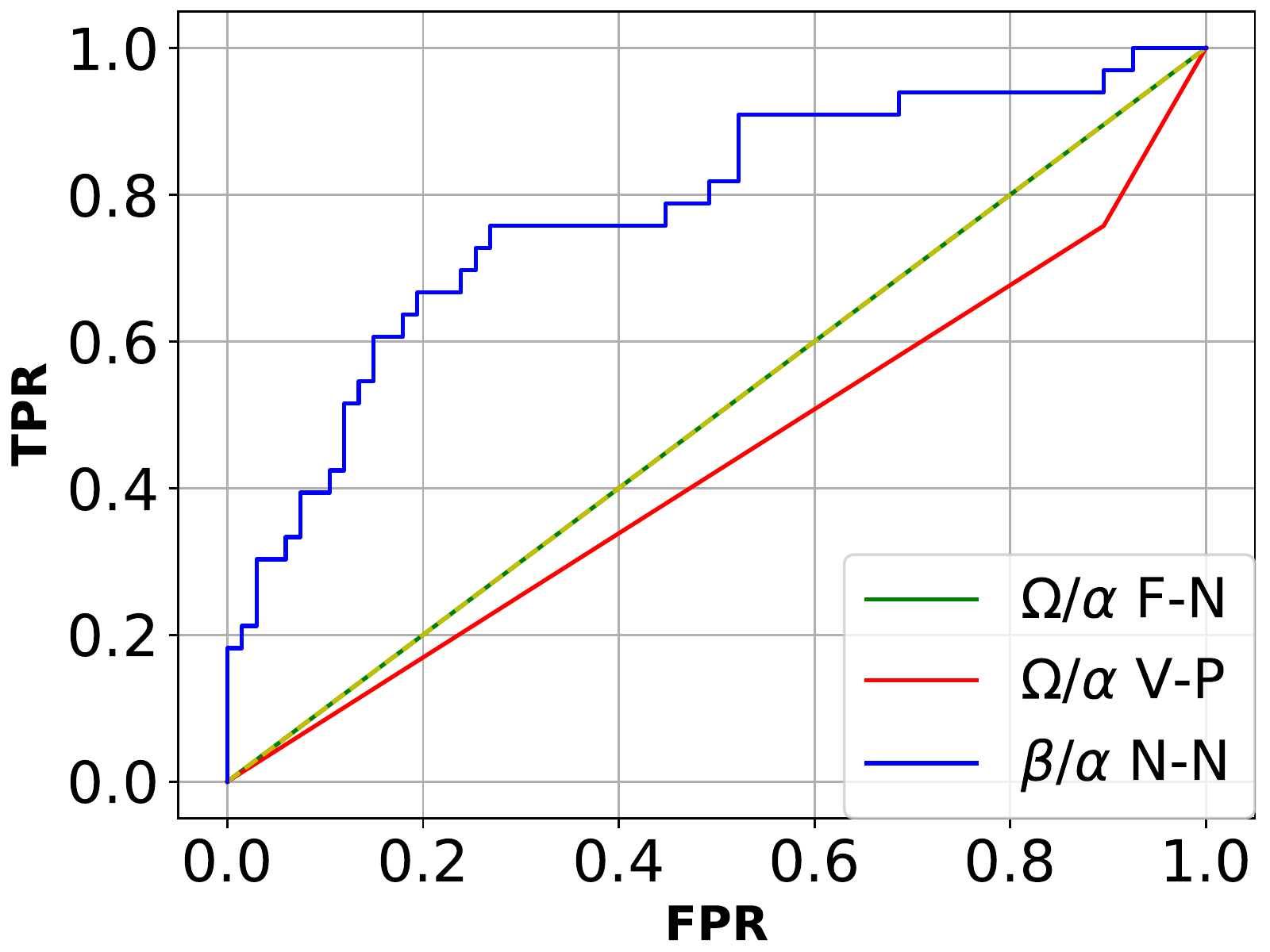}
		\caption{UV -- case 2}\label{fig:VUVL2}
			\vspace{0.1cm}
	\end{subfigure}
    \hfill
    \begin{subfigure}[b]{0.32\textwidth}
	\centering
		\includegraphics[width=1\textwidth,height=0.62\textwidth]{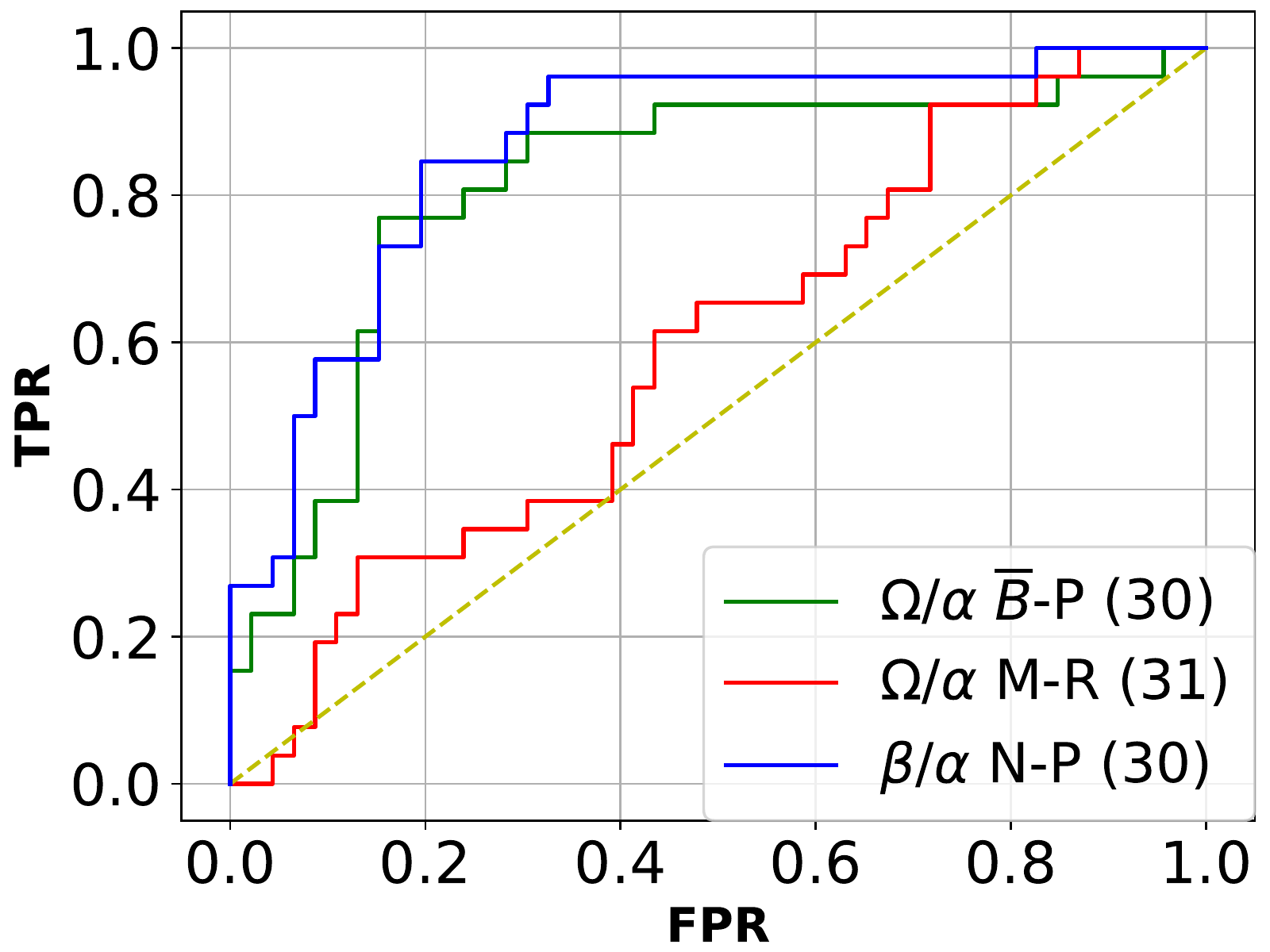}
		\caption{UV -- case 3}\label{fig:VUVL3}
			\vspace{0.1cm}
	\end{subfigure}
    \hfill
    \begin{subfigure}[b]{0.32\textwidth}
	\centering
		\includegraphics[width=1\textwidth,height=0.62\textwidth]{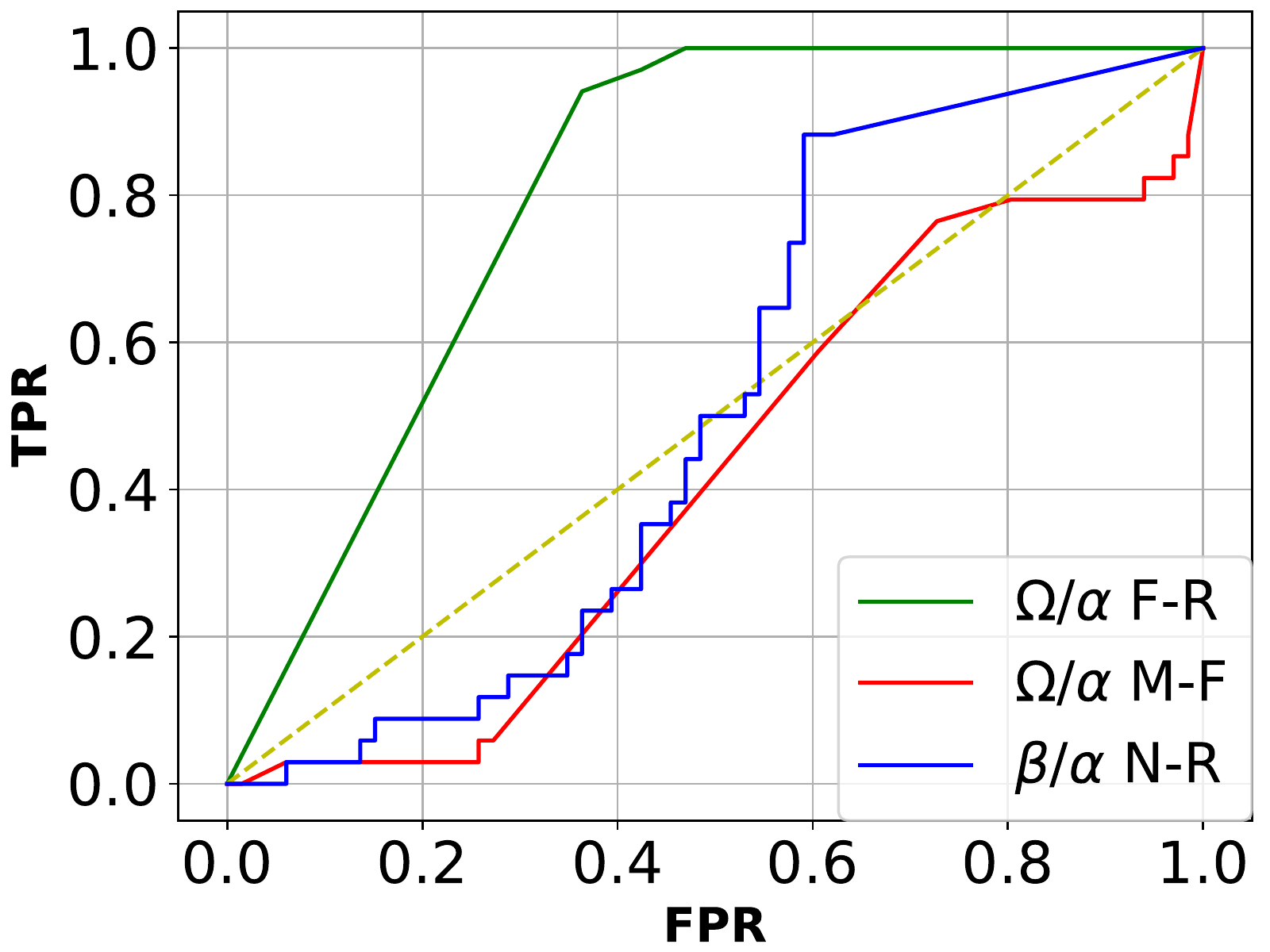}
		\caption{TM -- case 1}\label{fig:VTML1}
			\vspace{0.1cm}
	\end{subfigure}
    \hfill
    \begin{subfigure}[b]{0.32\textwidth}
	\centering
		\includegraphics[width=1\textwidth,height=0.62\textwidth]{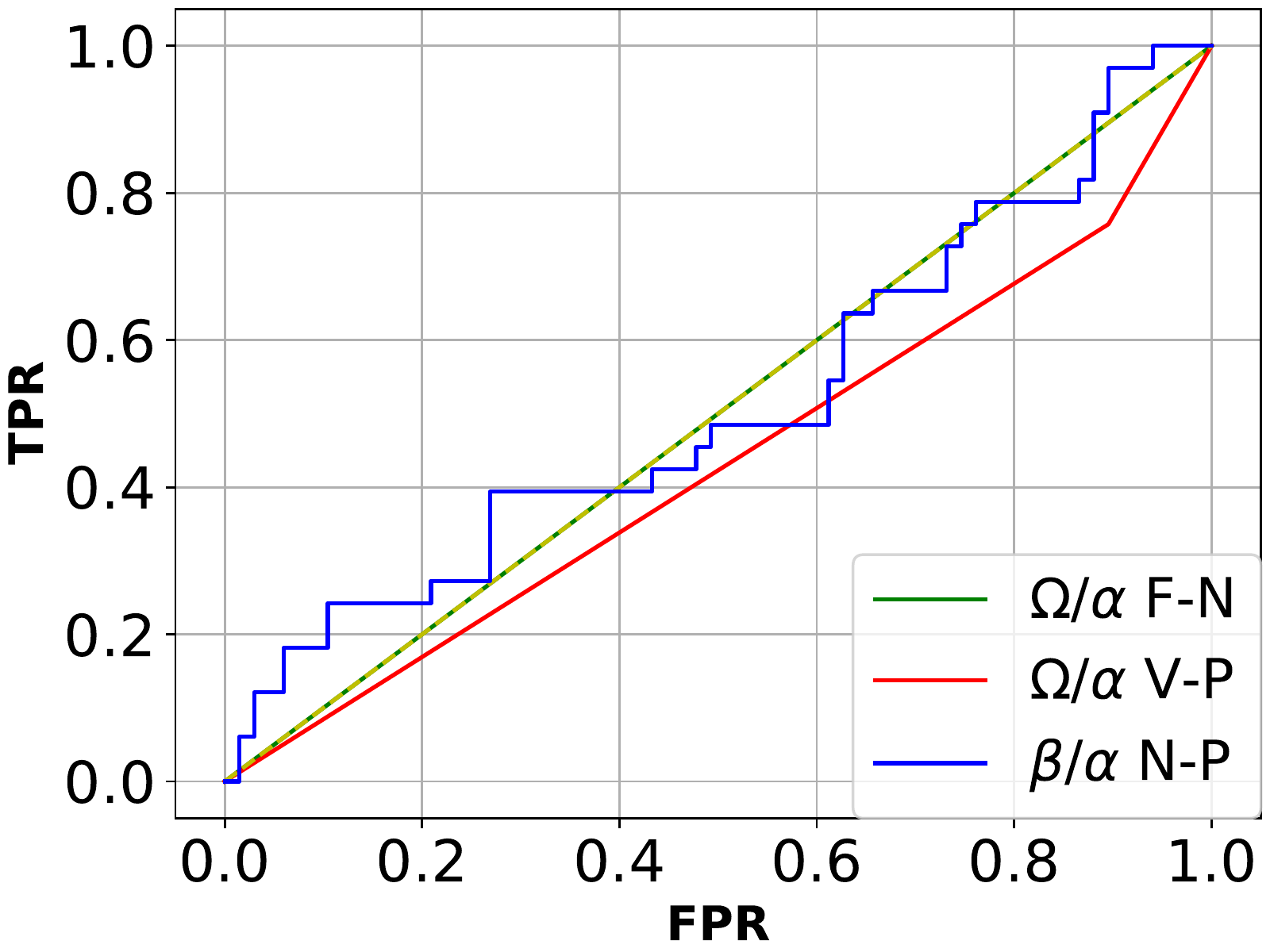}
		\caption{TM -- case 2}\label{fig:VTML2}
			\vspace{0.1cm}
	\end{subfigure}
    \hfill
    \begin{subfigure}[b]{0.32\textwidth}
	\centering
		\includegraphics[width=1\textwidth,height=0.62\textwidth]{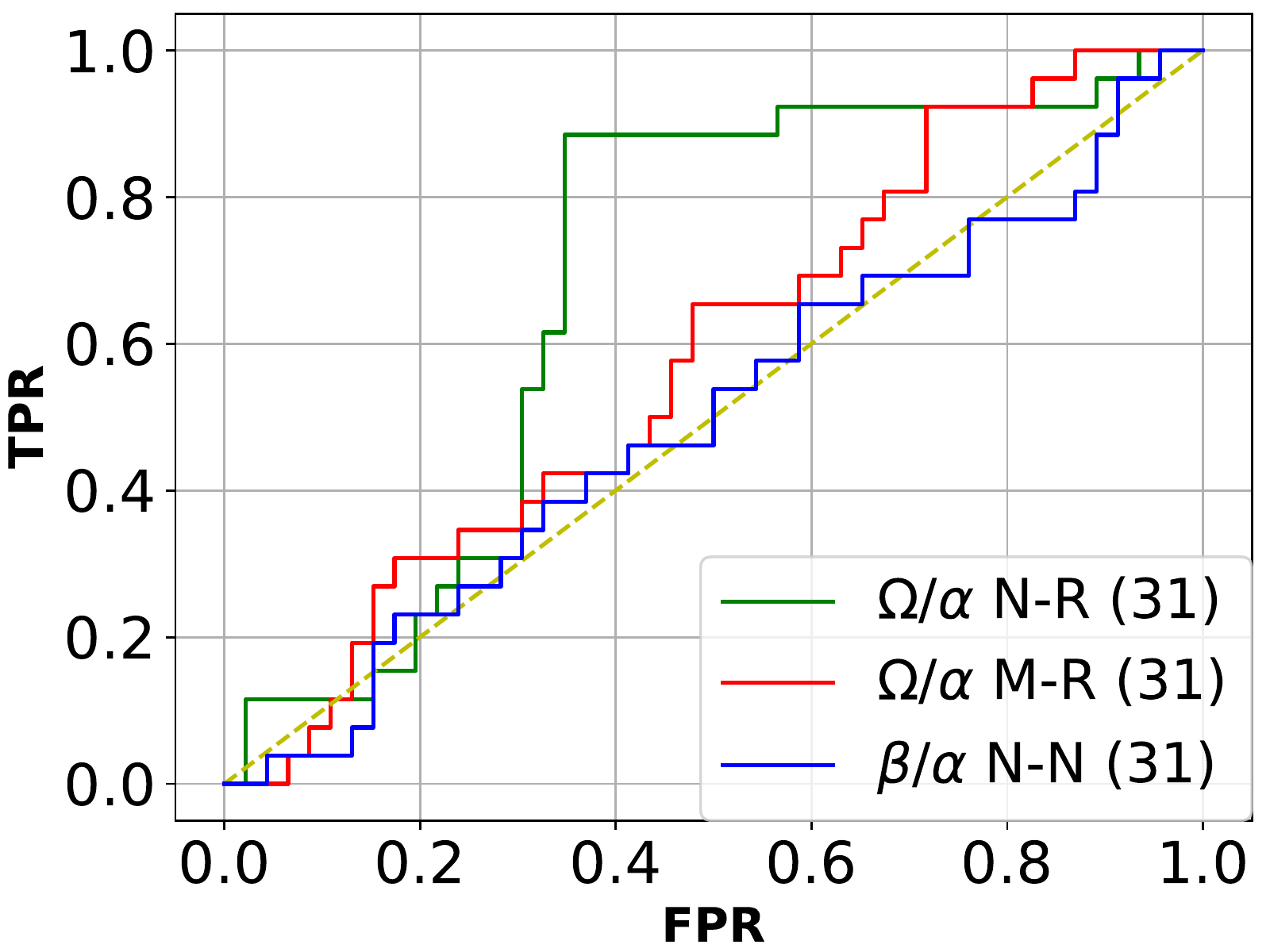}
		\caption{TM --case 3}\label{fig:VTML3}
			\vspace{0.1cm}
	\end{subfigure}
    \hfill
    \begin{subfigure}[b]{0.32\textwidth}
	\centering
		\includegraphics[width=1\textwidth,height=0.62\textwidth]{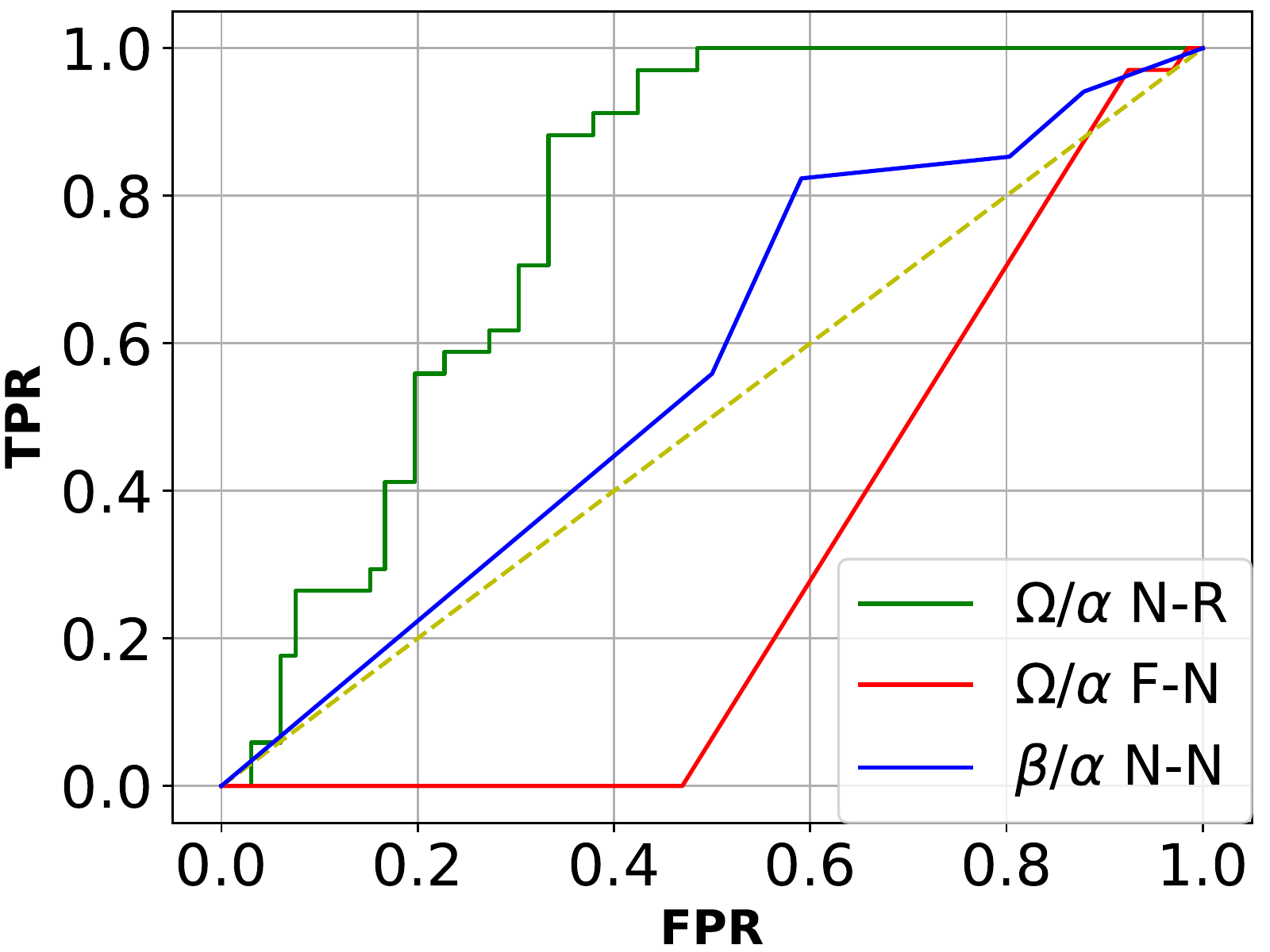}
		\caption{GP -- case 1}\label{fig:VGPL1}
			\vspace{0.1cm}
	\end{subfigure}
    \hfill
    \begin{subfigure}[b]{0.32\textwidth}
	\centering
		\includegraphics[width=1\textwidth,height=0.62\textwidth]{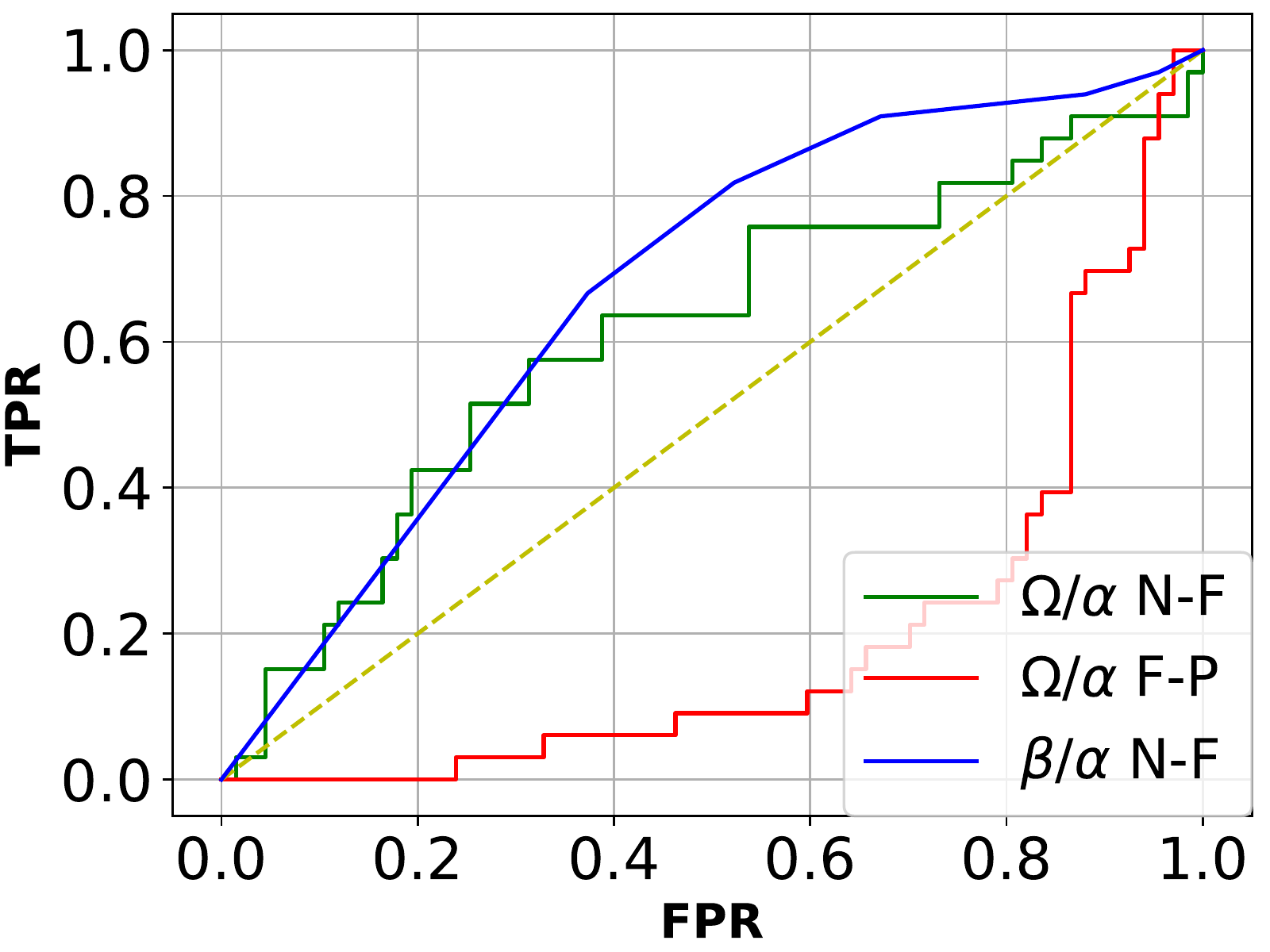}
		\caption{GP -- case 2}\label{fig:VGPL2}
			\vspace{0.1cm}
	\end{subfigure}
    \hfill
    \begin{subfigure}[b]{0.32\textwidth}
	\centering
		\includegraphics[width=1\textwidth,height=0.62\textwidth]{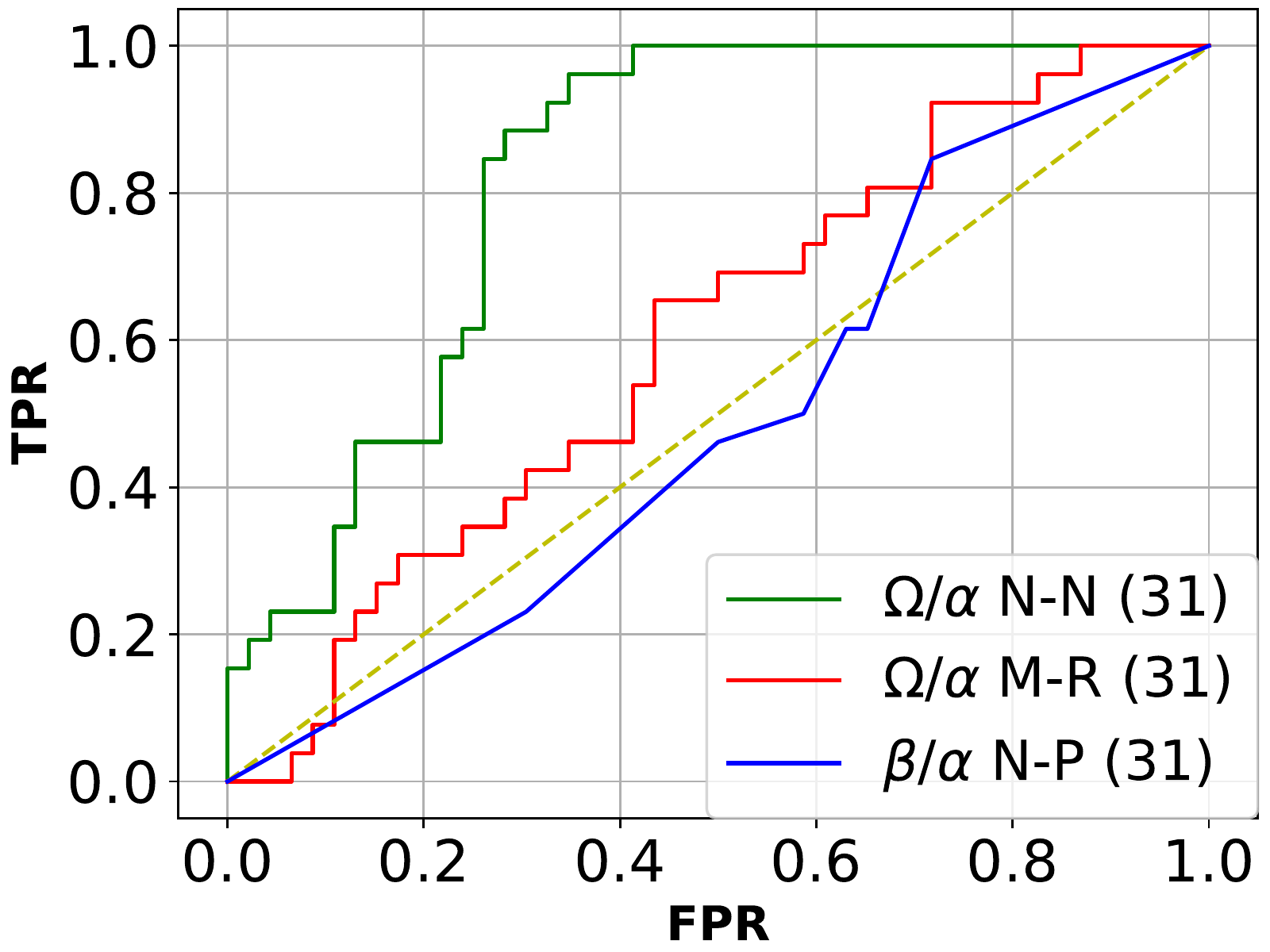}
		\caption{GP -- case 3}\label{fig:VGPL3}
			\vspace{0.1cm}
	\end{subfigure}
\caption{Selected ROC curves from validation data set per sensor and per case. Green curve belongs to the best ACC result of the cooperative decision-making, the red one to the worst cooperative result, and the blue one to the best local decision-making result.}\label{fig:ROC_V}
\end{figure*}

As a complement, figures \ref{fig:ROC_T} and \ref{fig:ROC_V} have the ROC curve per sensor and per case, local and cooperative, of the best results presented in the tables \ref{tab:Train} and \ref{tab:Validate}, in blue the local decision ($\beta$), and green the cooperative decision model ($\Omega$). These figures present the ROC curves per agent (each row) and per case (each column) where the resulting ROC curves are put together with the random result (yellow line) and the worst cooperative decision model per case (red line). Each cell has a box dialogue with the information related to the decision-making models drawn. 

\begin{figure*}
\centering
	\begin{subfigure}[b]{0.32\textwidth}
	\centering
		\includegraphics[width=1\textwidth,height=0.7\textwidth]{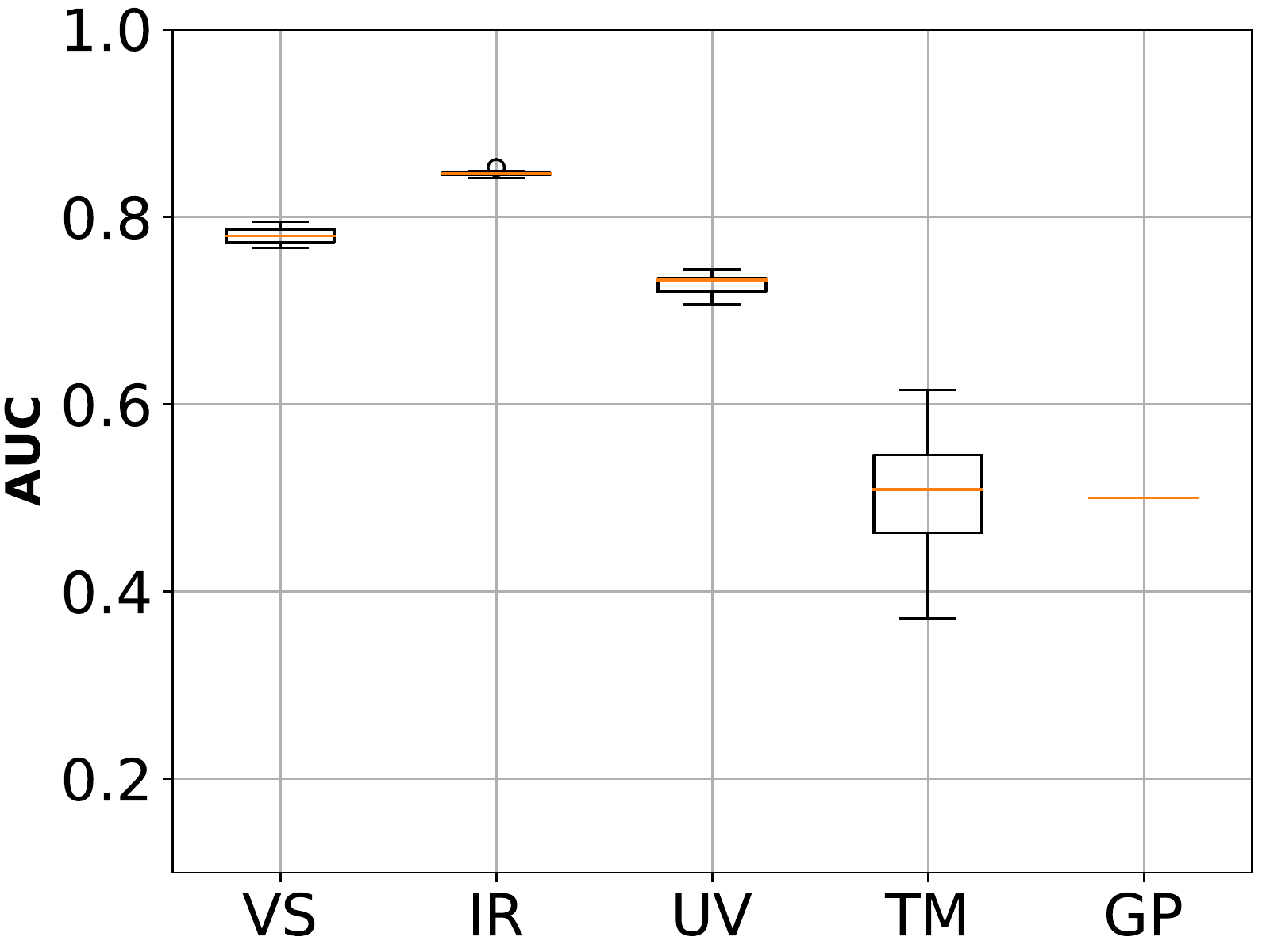}
		\caption{Case 1. Training results}\label{fig:ATLC1}
			\vspace{0.1cm}
	\end{subfigure}
    \hfill
    \hfill\begin{subfigure}[b]{0.32\textwidth}
	\centering
		\includegraphics[width=1\textwidth,height=0.7\textwidth]{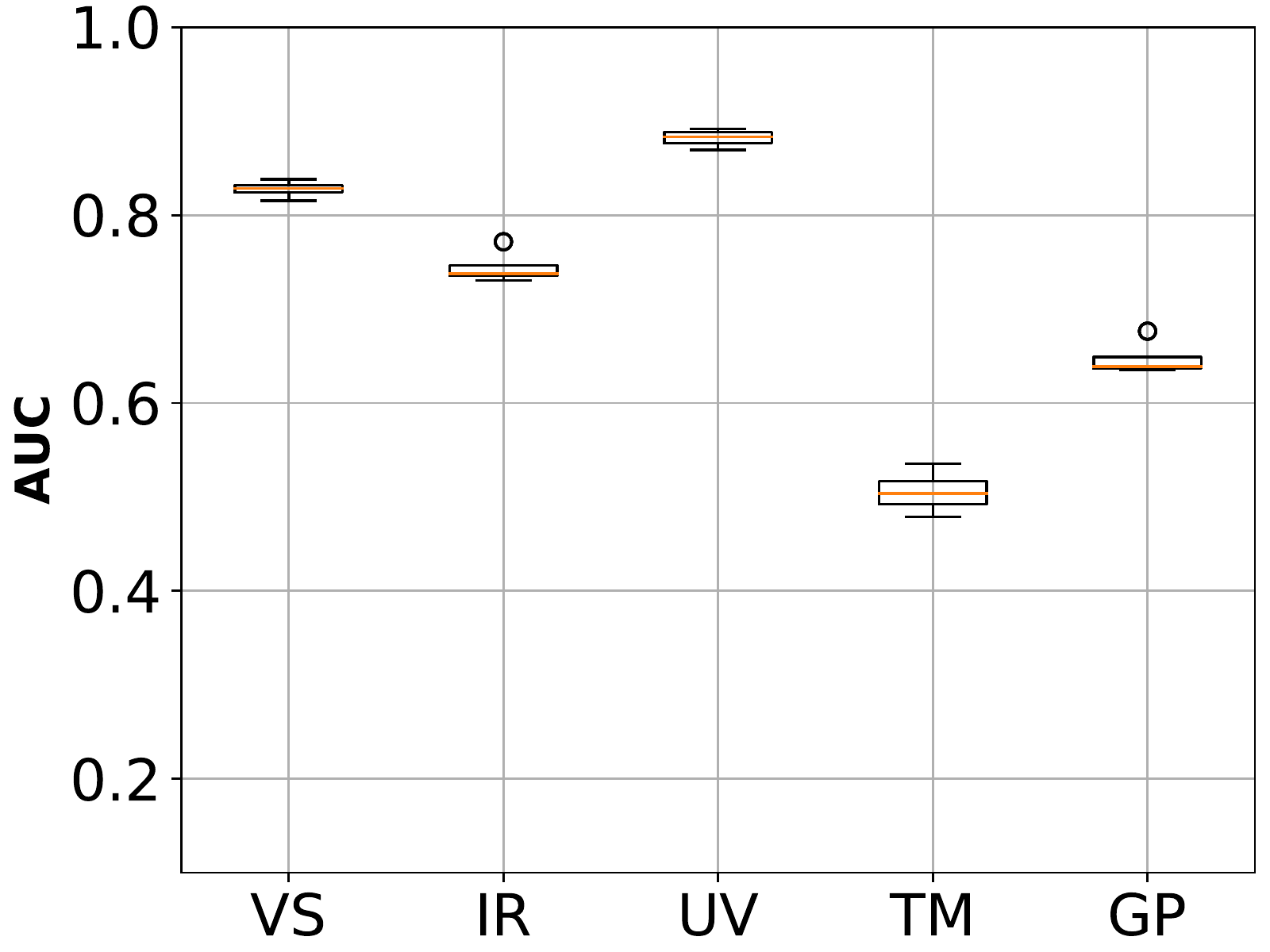}
		\caption{Case 2. Training results}\label{fig:ATLC2}
			\vspace{0.1cm}
	\end{subfigure}
    \hfill
    \begin{subfigure}[b]{0.32\textwidth}
	\centering
		\includegraphics[width=1\textwidth,height=0.7\textwidth]{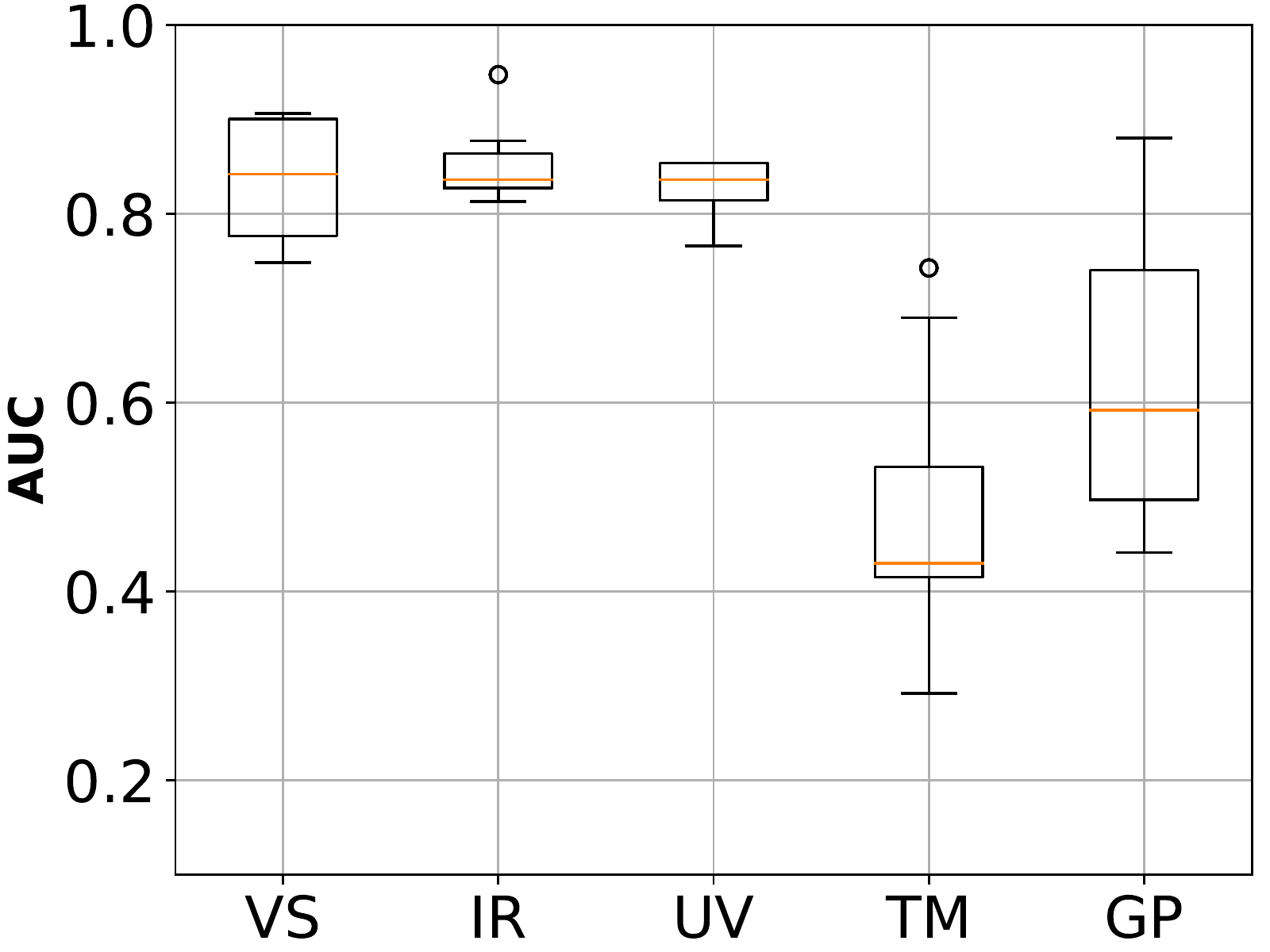}
		\caption{Case 3. Training results}\label{fig:ATLC3}
			\vspace{0.1cm}
	\end{subfigure}
    \hfill
    \begin{subfigure}[b]{0.32\textwidth}
	\centering
		\includegraphics[width=1\textwidth,height=0.7\textwidth]{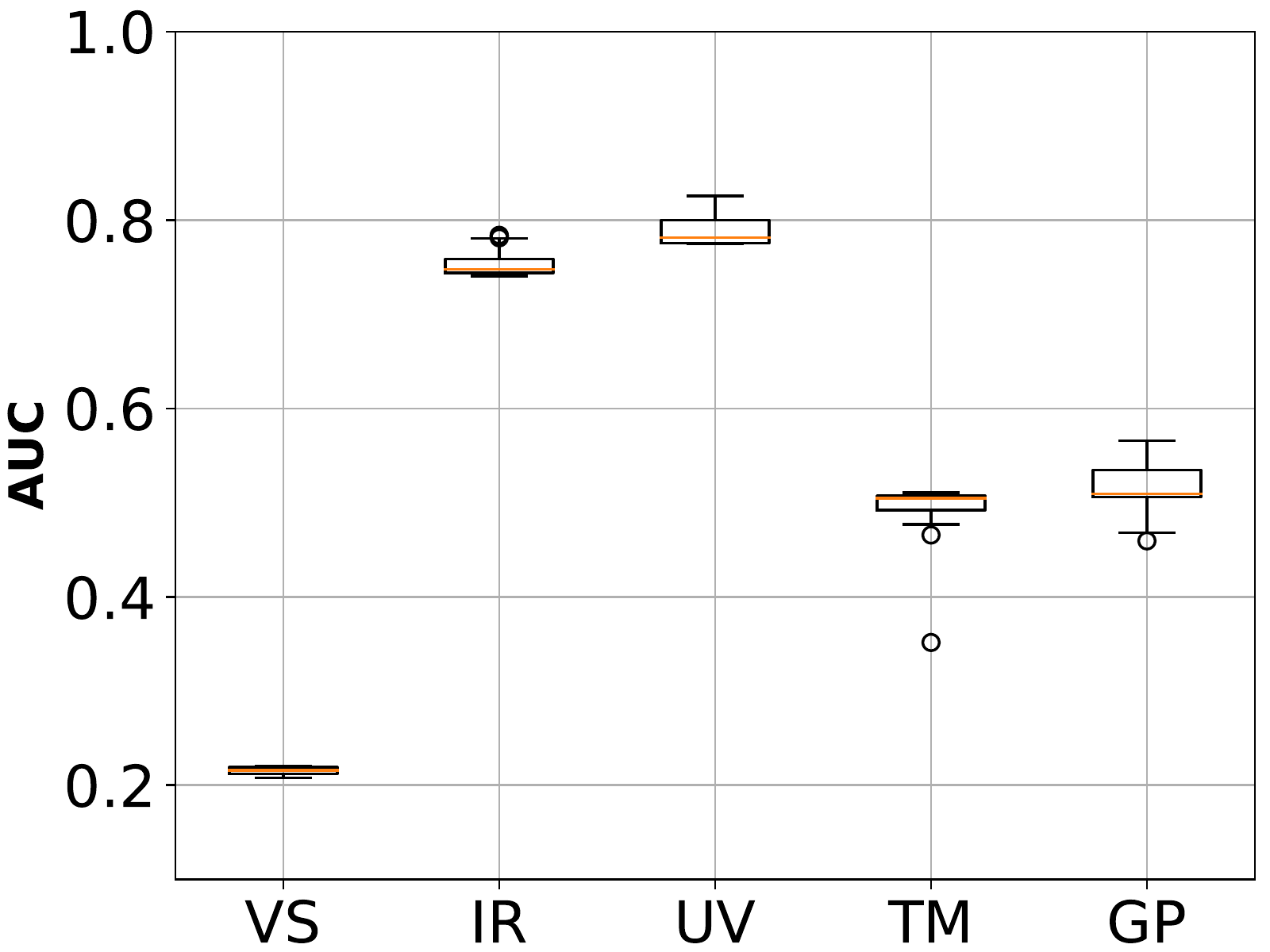}
		\caption{Case 1. Validation results}\label{fig:AVLC1}
			\vspace{0.1cm}
	\end{subfigure}
    \hfill
    \begin{subfigure}[b]{0.32\textwidth}
	\centering
		\includegraphics[width=1\textwidth,height=0.7\textwidth]{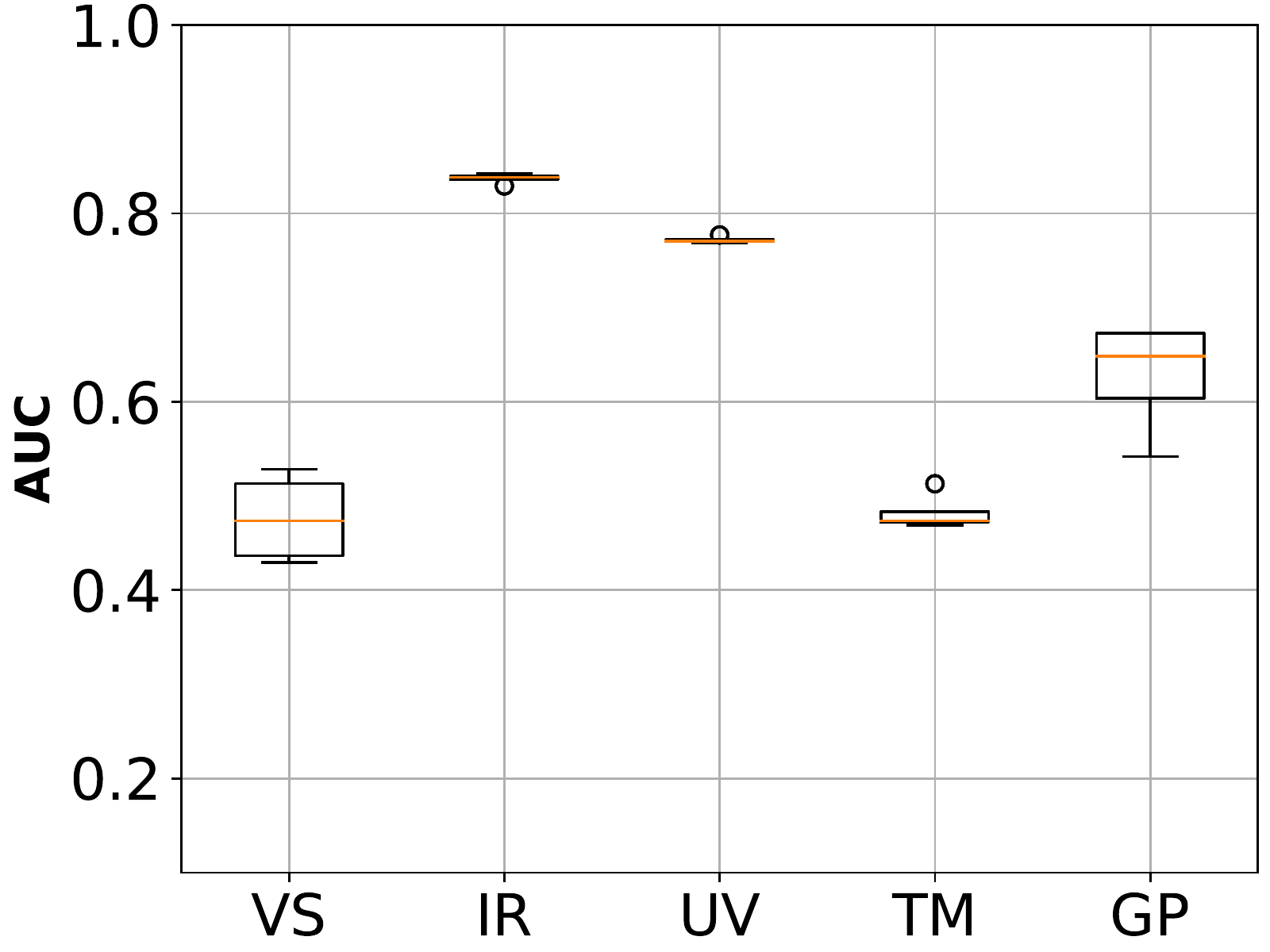}
		\caption{Case 2. Validation results}\label{fig:AVLC2}
			\vspace{0.1cm}
	\end{subfigure}
    \hfill
    \begin{subfigure}[b]{0.32\textwidth}
	\centering
		\includegraphics[width=1\textwidth,height=0.7\textwidth]{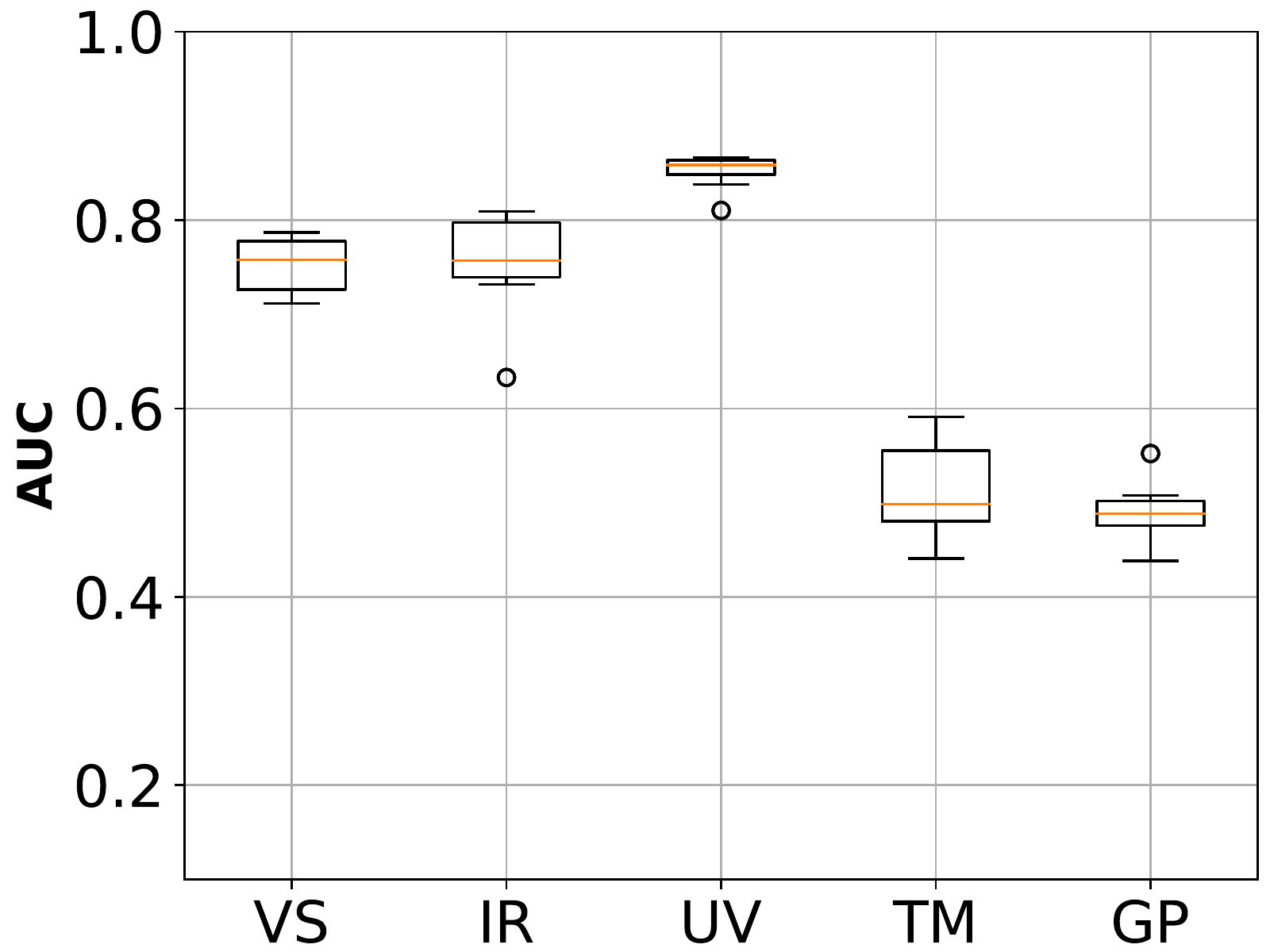}
		\caption{Case 3. Validation results}\label{fig:AVLC3}
			\vspace{0.1cm}
	\end{subfigure}
\caption{Distribution of the AUC for the local decision models per type of sensor and per stage.}\label{fig:AUC_L}
\end{figure*}

On the other hand, figures \ref{fig:AUC_L} and \ref{fig:AUC_C} have the box-plots of all the AUC results of the models evaluated, local ($\beta$) and cooperative ($\Omega$), respectively. In these figures, the columns are associated with the cases and the rows to the stages of model generation and evaluation (i.e. the first row shows the training stage while the second one the validation stage).

Figure \ref{fig:AUC_L} shows us that the information from the agent with a VS camera is the most affected by the distribution of the samples among training and validation. Since validation results of case one (figure \ref{fig:AVLC1}), show a performance decreasing for the AUC metric, in contrast to the performance of the same case for the training stage (figure \ref{fig:ATLC1}).
Likewise, figure \ref{fig:VSVL1} presents the low performance with the green curve which is close to the random case (yellow line).

Similar behaviour is detected for case two, however, the performance for the validation results in case two seems to have a lower affectation (figure \ref{fig:AVLC2} and figure \ref{fig:VSVL2}). Additionally, the results of case 3 indicate that the agent with the VS sensor keeps the performance of the AUC metric for both stages, training (figure \ref{fig:ATLC3}) and validation (figure \ref{fig:AVLC3}). It implies that this model is most robust to the illumination affectation, in contrast to cases 1 and 2.


\newpage
Moreover, figure \ref{fig:AUC_C} shows that the cooperative decision process increases and balance the AUC metric per agent since the integration of the local decisions contributes to the reduction of the uncertainty of the global decision. For the training stage, cases 1 and 2 have a similar distribution and mean value (orange line), the opposite happens for case three where the results are more scattered; the best results belong to case 3. A different result distribution can be seen at the validation stage, in which case 1 has the worst AUC values (figure \ref{fig:AVCC1}). Moreover, the AUC values distribution among cases 2 and 3 are similar. Nonetheless, the best achievement and mean values are given by case 3 (figures \ref{fig:ATCC3} and \ref{fig:AVCC3}). This case is the only one that has consistent performances of the AUC metric across data sets.


\section{Conclusion and Future Direction}\label{sec:conclusions}

Land mine detection is a task with many external constraints and challenges. This means that an effective detection model needs to avoid or reduce the sensor noise to achieve a robust decision. Our study of IEDs detection resulted in a system to integrate different sources of information; this study evaluates the performance of the decision-making models according to the training and validation set.

The distribution methodology of samples from the data set into training and validation is crucial when these samples are taken under the influence of varied external conditions; in this case, is particularly critical.

Looking at 3 cases considered, even though the performance in the training stage of the cases 1 and 2 are better than case 3, it is the validation stage determines the real performance of the models when they are tested with fresh data. The information recorded in the tables and figures in section \ref{SCM} indicates that samples distribution used for case 3 result in the best performance consistency between the training and validation stage. It means that the intelligent decision-making models from samples altered by different conditions are less sensible to sensor noise produced by external parameters.

An extension of this work is to acquire more samples with our CoD2M-MAPS with other environmental conditions (humidity and sunlight), terrains (sand and clay) and vegetation; to consider IEDs as disseminated along to the world. With these new samples, we can perform a new evaluation of the distribution of the samples to obtain the most robust decision-making system. On the other hand, the extension of the data set will allow evaluating and contrasting future IEDs detection systems.

\section*{Acknowledgement}

The ``Collaborative methodology for enhancing sustainability in rural communities and the use of land'' project \cite{project} received funds from De Montfort University (Leicester,UK),  Colciencias (Colombia, [grant number 647, 2014])  and Pontificia Universidad Javeriana (Bogot\'a, Colombia, [grant number VRI-05,2017]). 

\bibliographystyle{IEEEtran}
\bibliography{references}
\end{document}